\documentclass[10pt,twocolumn,letterpaper]{article}

\usepackage[pagenumbers]{cvpr} 

\definecolor{cvprblue}{rgb}{0.21,0.49,0.74}
\definecolor{highlight}{rgb}{0.90, 0.96, 1.00} 
\definecolor{highlightgray}{rgb}{0.88, 0.88, 0.88}
\usepackage[pagebackref,breaklinks,colorlinks,allcolors=cvprblue]{hyperref}
\definecolor{color1}{HTML}{edae49}
\definecolor{color2}{HTML}{d1495b}
\definecolor{color3}{HTML}{00798c}
\usepackage{natbib}
\usepackage{color}
\usepackage{graphicx}
\definecolor{DeepPink}{RGB}{255,20,147}
\usepackage{booktabs, multirow}
\usepackage{xcolor,amsmath}
\definecolor{darkblue}{rgb}{0,0.08,0.45}
\definecolor{cvprblue}{rgb}{0.21,0.49,0.74}
\definecolor{mygreen2}{RGB}{0 205 0}
\usepackage{colortbl}
\usepackage{amsfonts,bm,pifont}
\usepackage{wrapfig}
\usepackage{marvosym}
\usepackage{placeins}
\usepackage{tabularx} 
\usepackage{balance}
\usepackage{algorithm}
\usepackage{algpseudocode}
\def\paperID{9333} 
\def\confName{CVPR}
\def\confYear{2026}

\title{\raisebox{-1ex}{\includegraphics[height=3ex]{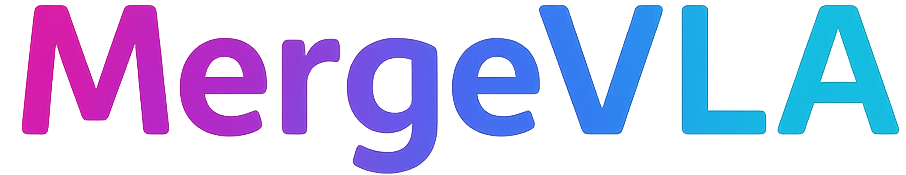}}: Cross-Skill Model Merging Toward a Generalist Vision-Language-Action Agent}
\author{
Yuxia Fu\thanks{The authors contribute equally to the research.} \quad 
Zhizhen Zhang\footnotemark[1] \quad
Yuqi Zhang \quad
Zijian Wang \quad
Zi Huang \quad
Yadan Luo \quad
\\
UQMM Lab, The University of Queensland\\ 
{\tt\small \{yuxia.fu, zhizhen.zhang, yuqi.zhang, zijian.wang, helen.huang, y.luo\}@uq.edu.au}
}

\begin{document}
\maketitle

\begin{abstract}
Recent Vision-Language-Action (VLA) models reformulate vision-language models by tuning them with millions of robotic demonstrations. While they perform well when fine-tuned for a single embodiment or task family, extending them to multi-skill settings remains challenging: directly merging VLA experts trained on different tasks results in near-zero success rates. This raises a fundamental question: what prevents VLAs from mastering multiple skills within one model? 
With an empirical decomposition of learnable parameters during VLA fine-tuning, we identify two key sources of \textit{non-mergeability}:
(1) Finetuning drives LoRA adapters in the VLM backbone toward divergent, task-specific directions beyond the capacity of existing merging methods to unify.
(2) Action experts develop inter-block dependencies through self-attention feedback, causing task information to spread across layers and preventing modular recombination.
To address these challenges, we present \textbf{MergeVLA}, a merging-oriented VLA architecture that preserves mergeability by design.
MergeVLA introduces sparsely activated LoRA adapters via task masks to retain consistent parameters and reduce irreconcilable conflicts in the VLM.
Its action expert replaces self-attention with cross-attention-only blocks to keep specialization localized and composable.
When the task is unknown, it uses a test-time task router to adaptively select the appropriate task mask and expert head from the initial observation, enabling unsupervised task inference.
Across LIBERO, LIBERO-Plus, RoboTwin, and multi-task experiments on the real SO101 robotic arm, MergeVLA achieves performance comparable to or even exceeding individually finetuned experts, demonstrating robust generalization across tasks, embodiments, and environments.
Project page: \url{https://mergevla.github.io/}
\end{abstract}

\section{Introduction}
\label{sec:intro}

Vision-Language-Action (VLA) models~\cite{OpenVLA,pi0,openvla-oft,pi0.5,rt1,rt2,gr00t,spatialvla,smolvla,octo,tinyvla,magma,vlm2vla} have recently enabled robot agents to perform complex manipulation tasks by fine-tuning large vision-language models (VLMs) with millions of robotic demonstrations. By reformulating action learning as a language generation or policy decoding problem, VLA models inherit broad visual grounding and semantic understanding from VLMs, and have shown notable performance in single-task or single-embodiment settings. However, real-world generalist agents must support \textit{multiple} skills, embodiments, and environments, requiring the ability to consolidate many independently fine-tuned VLA models into a single unified policy.

A natural approach is \textit{model merging}, which has proven effective in large language and vision models~\cite{TA,ties,iso-cts,wudi,dare,emr-merging,knots, robust-merging, pem-composition}. Those techniques can integrate multiple specialized models without joint retraining or revisiting their original datasets. Yet, when these approaches are applied to VLA experts fine-tuned on distinct manipulation tasks, the merged model exhibits \textbf{near-zero} success rate. This failure suggests that VLA finetuning induces structural specialization incompatible across tasks, which is rarely observed in conventional VLM merging.

To uncover the root causes, as shown in Figure~\ref{fig:comp_VLA}, we perform a fine-grained decomposition of trainable parameters across representative VLA architectures and identify two major sources of \emph{non-mergeability}:
First, LoRA updates within the VLM backbone diverge sharply across tasks.
Each manipulation task reshapes the pretrained representation space to satisfy its own perception-control alignment. Direct averaging ~\cite{TA,knots} or sign-resolved merging \cite{ties} thus reactivates irrelevant or even contradictory parameters, corrupting shared vision-language subspaces and degrading task-invariant semantics.
Second, the train-from-scratch action decoders accumulate strong task-specific dependencies across blocks through self-attention feedback. This coupling spreads localized task information globally, breaking modularity and preventing compositional merging even under identical architectures and initialization.

Built upon these findings, we introduce \textbf{MergeVLA}, a merge–oriented VLA architecture that preserves mergeability by design.
When executing a specific task, MergeVLA applies sparsely activated LoRA adapters, implemented via task masks, to selectively activate the merged parameters contributing to task-relevant responses while suppressing those that mislead other tasks.
Moreover, MergeVLA reconfigures the action expert to remove self-attention propagation, relying solely on cross-attention pathways. This eliminates the sustained accumulation of task dependence across blocks, allowing most layers to be effectively merged using simple weight averaging. Due to these architectural modifications, it surprisingly shows strong out-of-distribution (OOD) generalisation by 13.4\% higher success rate than VLA-Adapter under varying corruptions. However, the deeper blocks of the action expert, referred to as the \textit{expert head}, remain unmergeable due to their strong task specialization. Consequently, each task keeps its own expert head, which in practice is typically the final block $L$.
To address the challenge of \textit{mixed-task evaluation} \cite{knots,emr-merging}, where the task identity is unknown at inference time, we adopt a training-free \textit{test-time task router} that identifies the most relevant task directly from the input features. For each candidate task, the router runs the VLM with its corresponding task mask to obtain its hidden states, and then measures its response against principal components extracted from the value projections of a shared merged action expert. It selects the task with the highest response and activates the associated task mask and expert head. This allows MergeVLA to generalize to different tasks without additional supervision, enabling a single merged model to adaptively activate the right skill components for execution.

Extensive experiments show that MergeVLA achieves success rates of $\mathbf{90.2\%}$, $\mathbf{62.5\%}$, and $\mathbf{70.7\%}$ on the LIBERO~\cite{libero}, LIBERO-Plus~\cite{liberoplus}, and RoboTwin~\cite{robotwin} benchmarks under the \textit{mixed-task} evaluation setting, and $\mathbf{90.0\%}$ on real-world experiments with the SO101 robotic arm, demonstrating its effectiveness and robustness across cross-skill, cross-environment, and cross-embodiment evaluations. These results show that model merging is not only feasible for VLAs, but can serve as a scalable path toward generalist embodied agents.

\section{Related Work}
\label{sec:Related_Work}
\subsection{Vision-Language-Action Models}
Recent VLA models leverage the rich commonsense knowledge of large-scale VLMs and are fine-tuned on large-scale robotic trajectory data~\cite{bridgedatav2, OpenXEmbodiment,rt1,rt2} in an end-to-end manner to obtain more generalizable visuomotor policies~\cite{rt1,rt2,bridgedatav2,bridgedata, mtopt,qtopt,roboagent}. Among them, OpenVLA~\cite{OpenVLA} is a widely used open-source model based on Prismatic-7B~\cite{prismatic}, which discretizes robot actions into rarely used vocabulary tokens and generates them autoregressively. Similarly, $\pi_0$~\cite{pi0} builds on existing VLMs with a dual-system VLA architecture, introducing a lightweight action expert that uses conditional flow matching to generate continuous action chunks for faster inference.
Building on this, $\pi_{0.5}$~\cite{pi0.5} retains the text-generation ability of the VLM, enabling simultaneous high-level subtask descriptions and low-level action prediction. Recently, VLA-Adapter~\cite{vla-adapter} proposed a 0.5B-parameter dual-system model that coordinates the VLM and action expert through a novel mechanism, employing chunk-wise autoregressive generation for faster inference and strong performance at a small scale. While existing VLAs perform well on widely used robotic benchmarks~\cite{calvin, libero,robotwin,maniskill}, their multi-task ability relies on joint training~\cite{OpenVLA, pi0, pi0.5,vla-adapter}, making them inefficient. Besides, their tightly coupled dual systems~\cite{hirt, openhelix, dual1, pi0, vla-adapter} also hinder model merging. 

\begin{figure}[t]
    \centering
    \includegraphics[width=0.45\textwidth]{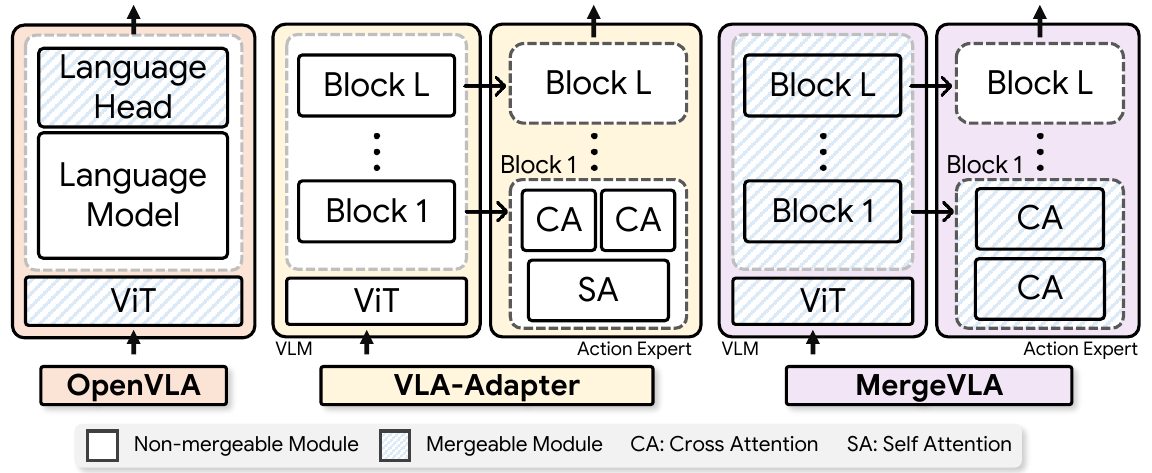}
    \caption{\textbf{Comparison between the structures of different VLAs.} OpenVLA uses a standard VLM for token-based action generation. VLA-Adapter adds an action expert with cross- and self-attention layers. MergeVLA simplifies this design by removing non-mergeable self-attention layers for effective merging.}
\label{fig:comp_VLA}
\vspace{-0.5cm}
\end{figure}

\begin{figure*}[t]
    \centering
    \includegraphics[width=\textwidth]{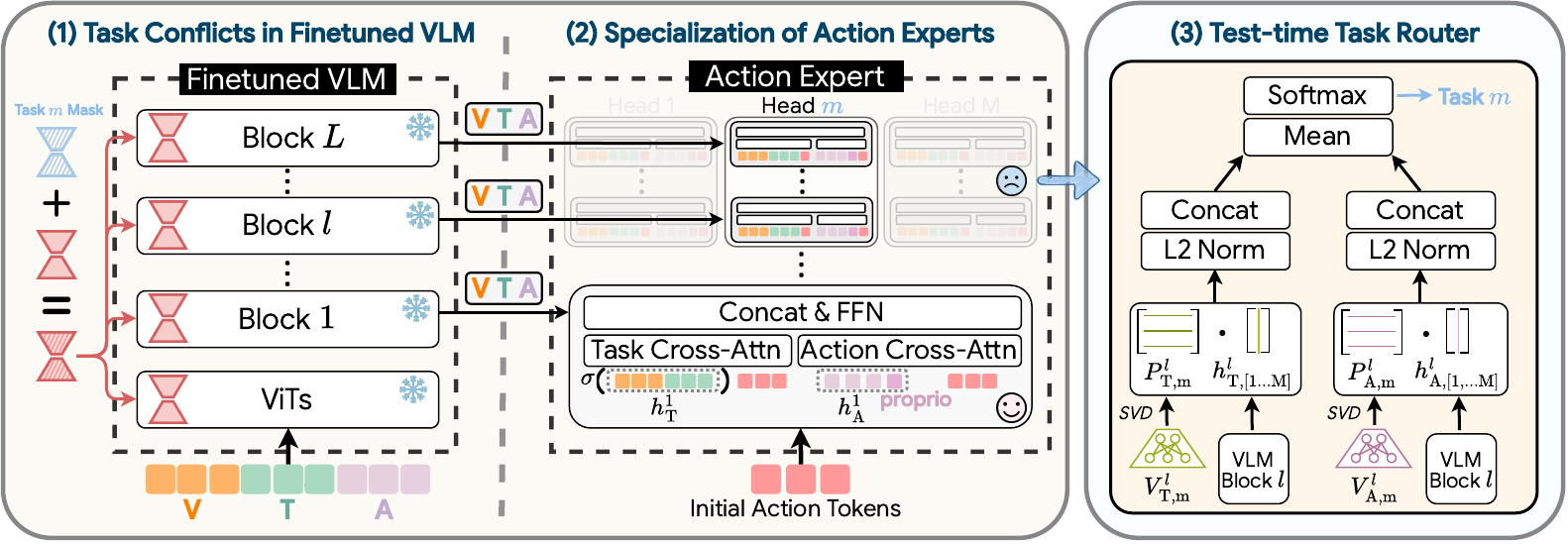}
    \caption{\textbf{Overview of MergeVLA architecture.} 
(1) To address destructive LoRA parameter interference in finetuned VLM, task masks are applied to all merged LoRA modules to selectively activate the merged parameters contributing to task-relevant responses while suppressing those that mislead other tasks.
(2) To solve the incompatibility of action experts, the architecture is redesigned to contain only cross-attention blocks and use $\mathrm{sigmoid}$ gate to preserve and rely on robust VLM features. Most blocks then can be merged except deeper blocks named \textbf{expert head} are left unmerged due to their task specification.
(3) To address the setting where the task identity is unknown at inference time, a training-free test-time task router is adopted to dynamically select task-specified components by computing task relevance from VLM hidden states in the value-based subspace of the merged action expert.
}
\label{fig:model_arch}
\end{figure*}

\subsection{Model Merging}
Model merging enables efficient knowledge reuse by combining existing model weights to construct a unified model capable of performing multiple tasks. Early studies~\cite{WA1,WA2,modelsoup,regmean,fishermerging} showed that simple weighted averaging of checkpoints can improve performance and introduce multi-task capability. Task Arithmetic (TA)~\cite{TA} treats the parameter difference between fine-tuned and pretrained models as a task vector and merges them to integrate task-specific knowledge. Nonetheless, it overlooks various conflicts that may arise among different task vectors. Subsequent methods~\cite{ties,dare,breadcrumbs,pcb-merging,cabs,cat-merging,wudi} handle conflicts via parameter pruning or rescaling, whereas subspace-based ones~\cite{tsvm,knots,iso-cts,twin-merging} apply low-rank decomposition (\textit{e.g.,} SVD) for more consistent merging. Other studies~\cite{pem-composition,robust-merging,knots,do-merging} design merging methods specifically for parameter-efficient fine-tuning (PEFT) modules. In contrast to these pre-merge approaches, methods like~\cite{twin-merging,emr-merging,calm-merging,tall-mask,smile} perform test-time merging to handle severe task interference, substantially improving performance. Yet, little work has explored model merging in VLA models. The recent ReVLA~\cite{revla} applies merging to gradually reverse the vision backbone to mitigate visual catastrophic forgetting and enhance domain generalization, rather than to enable multi-task VLA capabilities. To bridge this gap, we propose MergeVLA which enables lightweight multi-task robotic learning.

\section{Preliminary}
\label{sec:preliminary}

\noindent\textbf{Task Formulation.} We consider a collection of single-skill imitation learning datasets $\mathfrak{D}=\{\mathcal{D}_m\}^M_{m=1}$ where each dataset $\mathcal{D}_m$ corresponds to a distinct manipulation task. Each training set is denoted as $\mathcal{D}_m = \{\mathbf{I}^{v}_t, \mathbf{I}^{w}_t, L\}_{t=1}^{T}$, where $\mathbf{I}^{v}_t$ the third-person view image, $\mathbf{I}^{w}_t$ the wrist-mounted image, $L$ the task instruction at each time step $t$. Finetuning a pretrained VLA model on each dataset yields $M$ task-specific weights. The objective of model merging is to unify task-specific $\{\Theta_{1}, \ldots, \Theta_{M}\}$ into a single general agent $\Theta_{\operatorname{merge}}$ without retraining. 

\noindent\textbf{VLA Architectures.} Figure~\ref{fig:comp_VLA} illustrates the architectures of different VLAs. Some VLA models~\cite{OpenVLA, spatialvla, openvla-oft,tinyvla,magma} directly build upon existing VLMs, where only the language component is \textit{non-mergeable} as refered to our experiments in Table~\ref{tab:main_table_libero}. Others introduce an additional action expert~\cite{pi0,pi0.5,vla-adapter}, among which VLA-Adapter trains its action expert from scratch. However, this tightly coupled dual-system design makes the overall model difficult to merge. MergeVLA simplifies this structure by removing the non-mergeable self-attention layers, enabling all components except the expert head to be merged effectively.

\section{Our Approach: MergeVLA}
\label{sec: methods}
Finetuning VLA models on individual manipulation tasks, while effective, produces isolated experts that cannot be trivially merged. As discussed in Section~\ref{sec:libero}, applying standard model merging approaches to VLA specialists trained on LIBERO~\cite{libero} results in a complete breakdown, with all merged variants achieving \textbf{zero success rate}. This failure is unexpected given the success of merging strategies in language-only and vision-language domains, and suggests that VLA merging presents a \textit{more severe} challenge. To understand this phenomenon, we conduct a systematic analysis of the parameter space and architectural behavior of mainstream VLAs. Our findings, as depicted in Figure~\ref{fig:vis task interference}, show that VLA unmergeability arises from two complementary failure modes:
\begin{enumerate}
    \item \textbf{Destructive LoRA Parameter Interference}: As shown in the left plot, task-specific LoRA updates activate largely disjoint subsets of channels. When merging four tasks using TIES, the proportion of \textit{parameters that are relevant to exactly one task} (\textit{i.e.,} selfish parameters) already exceeds 75\%. Such extreme task exclusivity produces severe parameter conflicts, which directly cause naïve model-merging strategies to fail.
    \item \textbf{Architectural Incompatibility of Action Experts}: More critically, resolving LoRA interference is necessary but not sufficient. Even when the VLM is perfectly merged, simply averaging the action experts in architectures such as VLA-Adapter \cite{vla-adapter} still yields 0\% success. The right plot explains why: although the shallow blocks remain moderately aligned across tasks, the parameter distance \textit{explodes} in the final layers. Such divergence arises because the action expert is trained entirely from scratch and contains self-attention layers that accumulate task-specific differences over depth. Instead of providing modular transformations, these layers propagate and amplify task-dependent signals, causing deeper blocks to become pathologically specialized to individual tasks. As a result, their parameters are inherently \textit{irreconcilable} under any merging scheme.
\end{enumerate}
Additionally, prior merging studies \cite{knots,emr-merging} also note that the \emph{mixed-task} setting is considerably more challenging than per-task evaluation, where the task identity is unknown. In practice, many approaches rely on hand-picked priors, \textit{e.g.,} task identity or an expert-specific prompt, to perform well, and degrade rapidly without them. 
Motivated by these insights, we therefore structure \textbf{MergeVLA} as a unified framework that addresses these challenges by:  (1) Stabilizing VLM merging via task-specific masking (Sec \ref{sec:lora_merging}) to address extreme LoRA parameter conflict (\textcolor{color1}{\textbf{Q1}}); (2) redesigning the action expert (Sec \ref{sec:expert_merging}) to mitigate architectural incompatibility (\textcolor{color2}{\textbf{Q2}}); (3) introducing a learning-free task routing mechanism (Sec \ref{sec:routing}) to enable operation without a task identity at test time (\textcolor{color3}{\textbf{Q3}}). 

\subsection{Task Conflicts in LoRA-Finetuned VLM}\label{sec:lora_merging}
To address \textcolor{color1}{\textbf{Q1}}, we first consider a standard LoRA update.
Let $\Theta_0$ denote the pretrained checkpoint and $\Theta_m$ the LoRA-finetuned weights after finetuning on task 
$m \in \{1,\dots,M\}$.  
The task vector for task $m$ is defined as $\tau_m = \Theta_m - \Theta_0$.
Most data-free methods $\mathcal{R}(\cdot)$ 
construct a single merged update $\tau_{\mathrm{merge}}$ by merging the updates~\cite{TA,ties,emr-merging,tsvm,knots}:
\begin{equation}
\tau_{\mathrm{merge}} = \alpha\, \mathcal{R}(\{\tau_m\}_{m=1}^{M}), \quad
\Theta_{\mathrm{merge}} = \Theta_0 + \tau_{\mathrm{merge}},
\end{equation}
where $\alpha$ is the scaling factor. As $\tau_{\mathrm{merge}}$ is unusable directly due to the conflicts we identified, we must move beyond a single global update. We leverage a task-specific binary masking strategy $\mathbf{S}_m$ that isolates components in $\Theta_{\mathrm{merge}}$ beneficial to task $m$, while suppressing those encoding conflicts. Formally, 
\begin{equation}
\Theta_{\mathrm{merge}}^{(m)}
= \Theta_0 + \mathbf{S}_m \odot \tau_{\mathrm{merge}},
\end{equation}
where $\odot$ denotes element-wise multiplication. The mask $\mathbf{S}_m$ is constructed by a \textit{parameter-level consistency} test: A parameter is retained if and only if its task-specific $\tau_m$ is both (1) significant and (2) dominant over the scaled residual difference with $\tau_{\mathrm{merge}}$, indicating that it aligns with the overall merge and contributes positively to task $m$:
\begin{equation}
    \mathbf{S}_m = \mathbb{I}\left[\,|\tau_m| > \lambda\,|\tau_{\mathrm{merge}} - \tau_m|\,\right],
\end{equation}
where $\mathbb{I}[\cdot]$ is the indicator function that returns $1$ if the condition is true and $0$ otherwise, and mask ratio $\lambda$ controls the tolerance for disagreement. 
This principled approach of pruning parameters based on their alignment consistency, adapted here for LoRA merging, shares its core formulation with task-vector compression methods~\cite{tall-mask}.

\begin{figure}[t]
        \centering
        \begin{subfigure}[t]{.49\linewidth}
            \centering
            \includegraphics[width=1\linewidth]{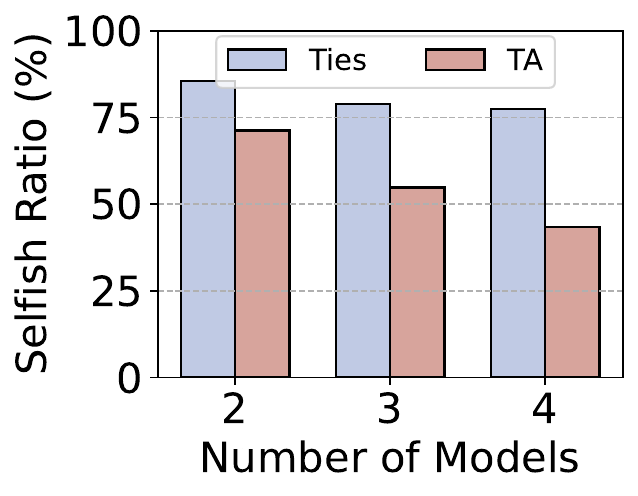}
        \end{subfigure}
        \begin{subfigure}[t]{.49\linewidth}
            \centering
            \includegraphics[width=1\linewidth]{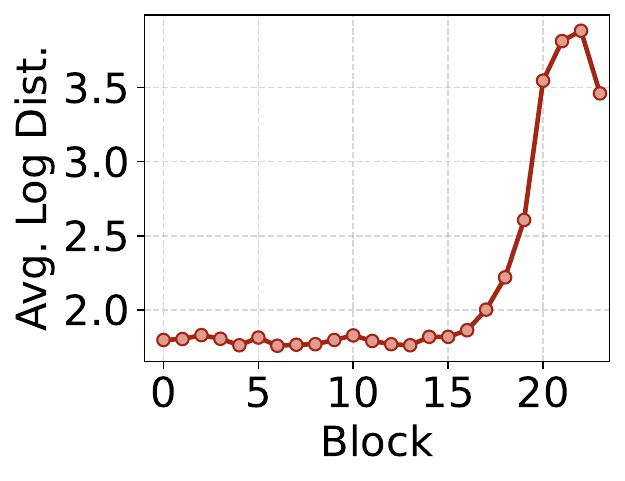}
        \end{subfigure}
        \caption{\textbf{Left:} Selfish ratio of the masks from TA \cite{TA} and TIES \cite{ties} by merging different numbers of tasks. The selfish ratio is computed following Equation~\ref{eq:selfish ratio}. \textbf{Right:} The average relative L2 distance across blocks between all pairs of action experts.}
        \label{fig:vis task interference}
\end{figure}

\noindent\textbf{Analysis of Parameter Selfishness.} Based on this formulation, we compute the proportion of \emph{selfish parameters}, namely those \textit{retained by exactly on}e task mask:
\begin{equation}
    \text{ratio}_{\mathrm{selfish}}
= \frac{1}{N}
\sum_{i=1}^{N} 
\mathbb{I}\!\left[\sum_{m=1}^{M} (\mathbf{S}_m)_i = 1\right],
\label{eq:selfish ratio}
\end{equation}
Empirically, as Figure~\ref{fig:vis task interference} (left) shows, this selfish ratio for both merging methods on LIBERO decreases as the number of merged tasks increases from 2 to 4.
In all cases with TIES, $\text{ratio}_{\mathrm{selfish}}$ rises at around 75\%, indicating that most parameters are selfish (kept exclusively by a single task) and underscoring the importance of task-specific masking to mitigate cross-task interference.
Notably, applying the task mask encourages some LoRA-finetuned parameters to revert toward their pretrained weights, improving merge stability.
A similar effect was observed in ReVLA~\cite{revla}, where VLA finetuning was shown to overwrite pretrained visual knowledge.
This forgetting becomes more pronounced when multiple experts are merged, as conflicting task updates distort the shared feature space.
By sparsely activating merged LoRA parameters through the proposed mask, MergeVLA preserves pretrained visual–language representations and mitigates cross-task conflicts, enabling more consistent and stable merging across tasks.

\subsection{Redesigning the Action Expert for Mergeability}\label{sec:expert_merging}
Our diagnosis proved that stabilizing the VLM (\textcolor{color1}{\textbf{Q1}}) is insufficient; the action expert itself is a fundamental barrier (\textcolor{color2}{\textbf{Q2}}). As shown in Section~\ref{sec:libero}, applying only the LoRA task mask still results in \textbf{0\% success rate}. Our analysis pinpoints the incompatibility with the VLA-Adapter~\cite{vla-adapter} architecture (Fig.~\ref{fig:comp_VLA}), which consists of $L$ transformer blocks trained from scratch. Each block contains a \textit{self-attention} layer on the block input $\mathbf{x}^i$, two \textit{cross-attention} layers conditioned on the VLM-provided hidden states $\mathbf{h}_\text{T}^i$ (task) and $\mathbf{h}_\text{A}^i$ (action), and a feed-forward network (FFN).  A gating function is applied to the task stream, \textit{i.e.}, $\hat{\mathbf{h}}_\text{T}^i = g(\mathbf{h}_\text{T}^i)$, with $g(\cdot)$ implemented as $\tanh$. 

Different from its flawed design, MergeVLA's redesign (Fig.~\ref{fig:model_arch}) is a principled response to nonmergeability diagnosis, introducing two key modifications:  
\begin{itemize}[leftmargin=*,noitemsep,topsep=3pt]
\item \textbf{Remove Self-Attention:} We eliminate the self-attention layers, retaining cross-attention \textit{only}. Since the expert is trained from scratch, self-attention layers develop strong, task-specific biases that are irreconcilable. Removing them forces the expert to rely on the robust, shared VLM features.
\item \textbf{Replace Gating:} We replace the $\tanh$ gate with a $\mathrm{sigmoid}$ gate. The original $\tanh$ gate can suppress VLM signals via negative activations, forcing the expert to rely on its own scratch-trained (and task-specific) parameters. $\mathrm{sigmoid}$ ensures VLM information is always preserved and balanced.
\end{itemize}
These two architectural changes alone are remarkably effective, achieving an \textit{13.4\%} higher success rate on the out-of-distribution (OOD) LIBERO-Plus \cite{liberoplus} testbed. This confirms our new design better leverages the VLM's \textit{robustness} and is inherently more \textit{generalizable}.

\noindent\textbf{Merging via Specialization Hierarchy.} Since the action expert is trained from scratch, existing task vector-based merging approaches are inapplicable here, as there is no shared initialization among different experts. Therefore, we adopt a simple weight-averaging strategy to merge the parameters of the action experts across tasks. We observe that such averaging works surprisingly well for the \textit{shallow} blocks of the action expert. However, it fails for the \textit{deeper} blocks, where the parameter discrepancy tasks increase sharply. 

This reflects strong task-specific specialization, and we refer to these divergent layers collectively as the \textbf{expert head}, denoted as $\mathbf{H}^{l\rightarrow L}$, spanning blocks $l$ through $L$. In most cases, $l = L$, meaning that only the final block requires separate handling.
We hypothesize that under regression-based training objectives, each expert head becomes \textit{highly specialized} to the action distribution of its corresponding task. 
Even \textit{small} discrepancies between these distributions can lead to incompatible weights, making simple parameter averaging ineffective. 
This effect is particularly pronounced in fine-grained manipulation tasks, where minor output deviations can cause entire trajectories to fail. 
Therefore, we leave the expert heads \textit{unmerged} and allow the model to use the corresponding head for each task.

\subsection{Test-time Task Routing}\label{sec:routing}
When the task identity is known, MergeVLA can already accomplish each skill by manually selecting the corresponding task mask $\mathbf{S}_m$ and expert head $\mathbf{H}_m^{l\rightarrow L}$. 
However, to operate without a known task identity (\textcolor{color3}{\textbf{Q3}}) at inference time, the model must dynamically select these components based solely on the input observations, enabling \textit{cross-skill} ability at a \textit{joint-task} level~\cite{knots}. 
To this end, we propose a test-time task routing mechanism that infers task relevance directly from the model's internal parameter subspaces, inspired by recent observations that fine-tuning pushes different tasks into distinguishable parameter subspaces~\cite{smile}. 
Given the merged LoRA parameters $\tau_{\mathrm{merge}}$ and task masks $\{\mathbf{S}_m\}_{m=1}^{M}$, 
we obtain $M$ masked VLM variants by applying each mask to the merged weights:
\begin{equation}
    \Theta^{(m)}_{\mathrm{merge}} = \Theta_0 + \mathbf{S}_m \odot \tau_{\mathrm{merge}}, \quad m = 1,\dots,M.
\end{equation}
Each masked VLM produces hidden states $\left[\mathbf{h}_\mathrm{T}^{l}, \mathbf{h}_\mathrm{A}^{l}\right]$ from block $l$, 
which is then forwarded to the corresponding $l$-th block of the action expert. Inside this block, two cross-attention paths are present: one conditioned on the task hidden state $\mathbf{h}_{\mathrm{T}}^{\,l}$ 
with parameters $(\mathbf{Q}_{\mathrm{T}}^{\,l},\mathbf{K}_{\mathrm{T}}^{\,l},\mathbf{V}_{\mathrm{T}}^{\,l})$, 
and the other conditioned on the action hidden state $\mathbf{h}_{\mathrm{A}}^{\,l}$ 
with $(\mathbf{Q}_{\mathrm{A}}^{\,l},\mathbf{K}_{\mathrm{A}}^{\,l},\mathbf{V}_{\mathrm{A}}^{\,l})$. 

A key design choice is \textit{which} subspace to use for routing. We hypothesize that query $\mathbf{Q}$ and $\mathbf{K}$ govern attentional selection, thus being sensitive to input scaling and risk collapsing into task-specific subspaces. Empirically, \textit{value-based} subspaces provide more stable and discriminative signals for routing, 
as they directly encode the task-dependent information written into the hidden states.
We therefore analyze the principal components of the value projection matrices $\textbf{V}$ of the $l$-th block for both paths via singular value decomposition (SVD):
\begin{equation}\label{eq:svd}
\begin{aligned}
\mathbf{V}_{\mathrm{T}}^{\,l} &= 
\mathbf{L}_{\mathrm{T}}^{\,l}\,\mathbf{\Sigma}_{\mathrm{T}}^{\,l}\,(\mathbf{R}_{\mathrm{T}}^{\,l})^\top,\\
\mathbf{V}_{\mathrm{A}}^{\,l} &= 
\mathbf{L}_{\mathrm{A}}^{\,l}\,\mathbf{\Sigma}_{\mathrm{A}}^{\,l}\,(\mathbf{R}_{\mathrm{A}}^{\,l})^\top.
\end{aligned}
\end{equation}
We retain the top-$k_r$ right singular vectors in each path to form the dominant content components of the expert subspace:
$\mathbf{P}_{\mathrm{T}}^{\,l} \in \mathbb{R}^{k_r\times d},$
$\mathbf{P}_{\mathrm{A}}^{\,l} \in \mathbb{R}^{k_r\times d},$
formed by taking the first $k_r$ rows of $(\mathbf{R}_{\mathrm{T}}^{\,l})^\top$ and $(\mathbf{R}_{\mathrm{A}}^{\,l})^\top$, respectively.    
For each task $m$, the hidden state from its masked VLM is projected onto these two subspaces to measure its \textit{activation strength},
\begin{equation}\label{eq:proj}
    r_{\mathrm{T},m} = \big\| \mathbf{P}_{\mathrm{T}}^{\,l} \mathbf{h}^{\,l}_{\mathrm{A},m} \big\|_2, r_{\mathrm{A},m} = \big\| \mathbf{P}_{\mathrm{A}}^{\,l} \mathbf{h}^{\,l}_{\mathrm{T},m} \big\|_2.
\end{equation}
Let $\mathbf{r}_{\mathrm{T}}, \mathbf{r}_{\mathrm{A}} \in \mathbb{R}^{M}$ collect the scores 
$\{r_{\mathrm{T},m}\}_{m=1}^{M}$ and $\{r_{\mathrm{A},m}\}_{m=1}^{M}$, respectively. 
Here, the combined score vector is given by $\mathbf{r} = \tfrac{1}{2}\big(\mathbf{r}_{\mathrm{T}} + \mathbf{r}_{\mathrm{A}}\big)$.
The routing probabilities are calculated by softmax $p_m = \operatorname{softmax}(\mathbf{r})_m.$
Then we can use $\arg\max_{m} \, p_m$ to choose the task index $m^{*}$. Once $m^{*}$ is determined, the model uses the corresponding task mask $\mathbf{S}_{m^{*}}$ 
and expert head $\text{H}^{l\rightarrow L}_{m^{*}}$ to perform the forward pass. 
This design allows the router to infer the underlying task purely from the input-driven hidden representations of the masked VLMs, without any additional training or supervision. 
In practice, we find that a \textit{single routing step} using the initial observation at $t{=}0$ is sufficient to identify the correct task. 
The selected task mask and expert head are then fixed for the rest of the episode, avoiding repeated routing during inference. 
Although this mechanism requires maintaining $M$ task masks and their corresponding action heads, the additional computational and parameter overhead is minimal. 

\section{Experiments}
\subsection{Experimental Setup}
\textbf{Simulation Benchmarks.} 
We evaluate MergeVLA on three simulation benchmarks: 

\noindent- \textbf{LIBERO}~\cite{libero} is a comprehensive benchmark consisting of four distinct task suites: Spatial, Object, Goal, and Long. Each suite contains 10 tasks with 50 demonstrations for every task. The benchmark enables the assessment of the robot’s multi-skill competence, including spatial relationships, object interactions, and task-specific objectives.

\begin{figure}[t]
    \centering
    \includegraphics[width=0.45\textwidth]{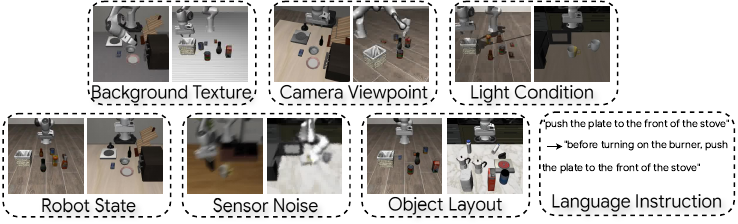}
    \caption{Seven perturbation types in the \textbf{LIBERO-Plus} benchmark, used to evaluate robustness under visual and language shifts.}
    \label{fig:env_libero-plus}
\end{figure}

\noindent- \textbf{LIBERO-Plus}~\cite{liberoplus} builds upon the original LIBERO benchmark by introducing seven distinct perturbations, as shown in Figure~\ref{fig:env_libero-plus}. The benchmark comprises 10,030 tasks, providing diverse settings for evaluating model generalization and robustness under shifts.

\begin{figure}[t]
    \centering
    \includegraphics[width=0.4\textwidth]{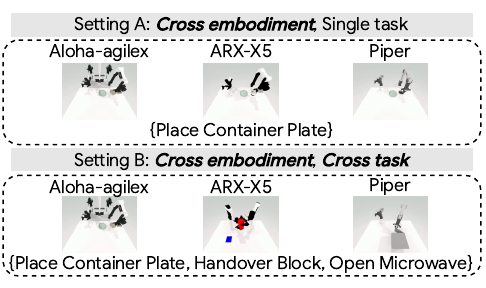}
    \caption{Experimental setup in the \textbf{RoboTwin} environment, featuring three robotic embodiments and a suite of manipulation tasks for cross-embodiment evaluation.}
    \label{fig:env_robotrwin}
\end{figure}

\noindent- \textbf{RoboTwin 2.0}~\cite{robotwin} serves as a cross-embodiment benchmark for dual-arm manipulation. As shown in Figure~\ref{fig:env_robotrwin}, we select three embodiments and four tasks for a comprehensive assessment of the cross-embodiment cross-skill ability.

\noindent\textbf{Real-World Robot Experiments.} For real-world evaluation, we deploy the SO101 robot within the LeRobot framework, as shown in Figure~\ref{fig:env_real}. 
We design three manipulation tasks: \textsc{cube picking}, \textsc{cube stacking}, and \textsc{cube pushing}. Each task includes 50 human-teleoperated demonstrations used for training. The detailed experimental setup is provided in Section~\ref{sec:real-world}.

\begin{table*}[t]
    \centering
    \small
    \caption{\textbf{LIBERO results across task splits.}
        Comparison between finetuned and merged variants of MergeVLA.
        All numbers are success rates (\%). $\mathbf{S}$ indicates that task masks are used during merging. ``Params (B)’’ denotes the total number of model parameters (in billions) required to evaluate on all four tasks, including the LLM backbone and the action expert. \textcolor{gray}{Gray-highlighted} rows correspond to per-task finetuned checkpoints evaluated on their own tasks, serving as upper-bound references for model merging.
    }
    \label{tab:main_table_libero}
    \begin{tabular}{lllcccccc}
    \toprule
    \textbf{Method} & \textbf{Merge Method} & \textbf{Merge Part} & \textbf{Params (B)} &
     \textbf{Spatial} & \textbf{Object} & \textbf{Goal} & \textbf{Long} & \textbf{Avg.} \\
    \midrule
    \multicolumn{9}{l}{\textbf{Single-task Finetuned Model}} \\
    \rowcolor{highlightgray}
    $\mathrm{OpenVLA}$~\cite{OpenVLA} & - & - & 7 $\times$ 4  & 84.7 & 88.4 & 79.2 & 53.7 & 76.5 \\
    \rowcolor{highlightgray}
    $\mathrm{VLA}$-$\mathrm{Adapter}$~\cite{vla-adapter} & - & - & 0.68 $\times$ 4 & 99.6& 99.6 & 98.2& 96.4 & 98.5\\
    \rowcolor{highlightgray}
      $\mathrm{MergeVLA}$ & - & - & 0.68 $\times$ 4 & 98.0 & 98.6 & 95.0 & 95.0 & 96.7 \\
    \midrule
    \multicolumn{9}{l}{\textbf{Merged Model}} \\
    $\mathrm{OpenVLA}$ & TA~\cite{TA} & Vision Backbones  & 7 $\times$ 4 & 56.6 & 58.0 & 55.6 & 6.6 & 44.2 \\
    $\mathrm{OpenVLA}$ & TA~\cite{TA} & All & 7 & \textcolor{gray}{0.0} & \textcolor{gray}{0.0} & \textcolor{gray}{0.0} & \textcolor{gray}{0.0} & \textcolor{gray}{0.0} \\
    $\mathrm{OpenVLA}$ & TA~\cite{TA} + $\mathbf{S}$ & All & 7 & 74.2 & 82.6 & 68.8 & 24.0 &  62.4\\
    $\mathrm{VLA}$-$\mathrm{Adapter}$ & TA~\cite{TA} & All  & 0.68 & \textcolor{gray}{0.0} & \textcolor{gray}{0.0} & \textcolor{gray}{0.0} & \textcolor{gray}{0.0} & \textcolor{gray}{0.0} \\
    $\mathrm{VLA}$-$\mathrm{Adapter}$ & TA~\cite{TA} + $\mathbf{S}$ & All & 0.68 & \textcolor{gray}{0.0} & \textcolor{gray}{0.0} & \textcolor{gray}{0.0} & \textcolor{gray}{0.0} & \textcolor{gray}{0.0} \\
    $\mathrm{VLA}$-$\mathrm{Adapter}$ & TA~\cite{TA} + $\mathbf{S}$ & Except $\mathbf{H}^{L\rightarrow L}$  & 0.70 & 50.2 & 34.6 & 0.0 & 7.4 & 23.1 \\
     \rowcolor{highlight}
    $\mathrm{MergeVLA}_\mathrm{EMR}$ & EMR~\cite{emr-merging} & & & 96.0 & 63.2 & 62.0 & 40.6 & 65.5 \\
     \rowcolor{highlight}
    $\mathrm{MergeVLA_{TSV}}$ & TSV~\cite{tsvm} + $\mathbf{S}$ & & & \textbf{99.4} & 97.8 & 74.4 & 54.8 & 81.6 \\
     \rowcolor{highlight}
    $\mathrm{MergeVLA_{KnOTS}}$ & KnOTS\cite{knots} + $\mathbf{S}$ & & & 96.8 & 98.8 & 84.8 & 71.4 & 88.0 \\
     \rowcolor{highlight}
     $\mathrm{MergeVLA_{TA}}$ & TA~\cite{TA} + $\mathbf{S}$ & Except $\mathbf{H}^{L\rightarrow L}$ & 0.70 & 98.0 & \textbf{98.8} & 85.4 & 76.6 & 89.7 \\
     \rowcolor{highlight}
    $\mathrm{MergeVLA_{WUDI}}$ & WUDI~\cite{wudi} + $\mathbf{S}$  & & & 97.6 & 98.2 & 85.6 & 78.2 & 89.9 \\
    \rowcolor{highlight}
    $\mathrm{MergeVLA_{TIES}}$ & TIES~\cite{ties} + $\mathbf{S}$ & &  & 94.8 & 94.6 & \textbf{91.8} & \textbf{79.4} & \textbf{90.2}  \\
    \bottomrule
    \end{tabular}
\end{table*}

\noindent\textbf{Implementation Details.} 
Our vision-language backbone is Qwen2.5-0.5B~\cite{qwen25}. By default, we set $l=L$, $k_r=8$, mask ratio $\lambda=0.6$, merging scaling factor $\alpha=1$.
All finetuning are conducted on a single NVIDIA A6000 Ada GPU (48\,GB). Other hyperparameters can be found in Appendix.

\subsection{Results on LIBERO}
\label{sec:libero}
We first reproduce results of OpenVLA~\cite{OpenVLA} and VLA-Adapter~\cite{vla-adapter} on four tasks, as highlighted in grey in Table~\ref{tab:main_table_libero}. OpenVLA performs notably worse than the others, with an average success rate of 76.5\% and only 53.7\% on the long-horizon LIBERO-Long task. Our MergeVLA, modified from VLA-Adapter to be merge-friendly, achieves comparable fine-tuning performance, indicating that our structural changes preserve the original capability. 

We have tried evaluating single-task finetuned models on unseen tasks (\textit{e.g.,} testing a Spatial expert on the Object suite), all methods achieve \emph{0\%} success across tasks.
This clearly indicates that existing VLA models lack cross-skill generalization.
The lower part of Table~\ref{tab:main_table_libero} reports the merging results. Directly applying TA~\cite{TA} to OpenVLA fails entirely, while merging only the vision and projector components improves to 44.2\%. Adding a task-specific mask further raises the average success rate to 62.4\%, though performance remains unbalanced. On VLA-Adapter, tight coupling between the VLM and action expert makes merging difficult—even with masking, unless the final action block is excluded. As shown in the blue-highlighted rows, our MergeVLA consistently outperforms other merging baselines. With TIES~\cite{ties} and WUDI~\cite{wudi}, it achieves balanced results across all four tasks (up to 90.2\% average success rate), while remaining lightweight and only 6.5\% below fine-tuning performance. These results confirm that MergeVLA enables efficient knowledge reuse and strong multi-task performance with a compact VLA model.

\begin{table*}[t]
\centering
\small
\caption{\textbf{Robustness of different models under visual and language shifts on LIBERO-Plus.}
All results are success rates (\%) averaged over 4 task suites. \textcolor{gray}{Gray-highlighted} rows correspond to per-task finetuned checkpoints evaluated on their own tasks, serving as upper-bound references for model merging. \textbf{Shift definitions:}
S1 – Background Textures; 
S2 – Camera Viewpoints; 
S3 – Language Instructions; 
S4 – Lighting Conditions; 
S5 – Object Layout; 
S6 – Robot States; 
S7 – Sensor Noise.}
\vspace{5pt}
\begin{tabular}{
    l
    *{8}{>{\centering\arraybackslash}p{1.25cm}}
}
\toprule
\textbf{Method} & \textbf{S1} & \textbf{S2} & \textbf{S3} & \textbf{S4} & \textbf{S5} & \textbf{S6} & \textbf{S7} & \textbf{Avg.} \\
\midrule
\multicolumn{9}{l}{\textbf{Single-task Finetuned Model}} \\
\rowcolor{highlightgray}
$\mathrm{OpenVLA}$~\cite{OpenVLA} & 34.8 & 0.8 & 23.0 & 8.1 & 28.5 & 3.5 & 15.2 & 16.3 \\
  \rowcolor{highlightgray}
$\mathrm{\pi_0}$~\cite{pi0} & 81.4 & 13.8 & 58.8 & 85.0 & 68.9 & 6.9 & \textbf{79.0} & 56.3 \\
   \rowcolor{highlightgray}
$\mathrm{VLA}$-$\mathrm{Adapter}$~\cite{vla-adapter} & 76.6 & 36.4 & 73.8 & 71.0 & 70.2 & 37.4 & 57.2 & 59.0 \\
 \rowcolor{highlightgray}
$\mathrm{MergeVLA}$ & 92.7 & \textbf{62.4} & 75.7 & \textbf{92.7} & 73.7 & \textbf{46.4} & 74.7 & \textbf{72.4} \\
\midrule
\multicolumn{9}{l}{\textbf{Merged Model}} \\
$\mathrm{VLA}$-$\mathrm{Adapter}$~\cite{vla-adapter} &  15.7 & 6.6 & 17.6 & 11.2 & 15.0 & 4.1 & 7.1 & 10.8 \\
 \rowcolor{highlight}
$\mathrm{MergeVLA}_{\mathrm{TSV}}$ & 64.3 & 44.8 & 58.8 & 72.9 & 59.4 & 30.4 & 52.7 & 53.5 \\
\rowcolor{highlight}
$\mathrm{MergeVLA}_{\mathrm{TA}}$ & 78.2 & \textbf{53.1} & \textbf{68.4} & 79.0 & 65.0 & \textbf{34.6} & 62.7 & 61.6 \\
 \rowcolor{highlight}
$\mathrm{MergeVLA}_{\mathrm{TIES}}$ & \textbf{85.7} & 50.7 & 66.0 & \textbf{84.2} & \textbf{68.1} & 30.3 & \textbf{66.0} & \textbf{62.5} \\
\bottomrule
\end{tabular}
\label{tab:robustness}
\end{table*}

\begin{table}[t]
    \centering
    \small
    \caption{\textbf{RoboTwin success rates (\%) of different variants of MergeVLA across embodiments and tasks.} $\mathbf{T}_1$: \textsc{Place Container Plate}, $\mathbf{T}_2$: \textsc{Handover Block}, $\mathbf{T}_3$: \textsc{Open Microwave}. \textcolor{gray}{Gray-highlighted} rows correspond to per-task finetuned checkpoints evaluated on their own tasks, serving as upper-bound references for model merging.}
    \begin{tabularx}{\columnwidth}{lcccX}
    \toprule
    \multicolumn{5}{c}{\textit{Setting A: \textbf{Cross embodiments}, Single task}} \\
    \midrule
    \textbf{Method} &  \multicolumn{3}{c}{$\mathbf{T}_1$} & \multirow{2}{*}{Avg.} \\
     & Aloha & ARX & Piper &  \\
    \rowcolor{highlightgray}
    Single-task Finetuned & 90.0 & 90.0 & 84.0 & 88.0\\
    \rowcolor{highlight}
    $\mathrm{MergeVLA}_{\mathrm{TA},\mathbf{H}^{(L-1)\rightarrow L}}$ & 86.0 & 82.0 & 68.0 & 78.7\\
    \rowcolor{highlight}
    $\mathrm{MergeVLA}_{\mathrm{TIES},\mathbf{H}^{(L-1)\rightarrow L}}$ & \textbf{88.0} & \textbf{92.0} & \textbf{86.0} & \textbf{88.7}\\
    \midrule
    \multicolumn{5}{c}{\textit{Setting B: \textbf{Cross embodiments, Cross task}}} \\
    \midrule
    \textbf{Method} & $\mathbf{T}_1$ & $\mathbf{T}_2$ & $\mathbf{T}_3$ & \multirow{2}{*}{Avg.} \\
     & Aloha & ARX & Piper & \\
    \rowcolor{highlightgray}
    Single-task Finetuned & 90.0 & 46.0 & 92.0 & 76.0 \\
    \rowcolor{highlight}
    $\mathrm{MergeVLA}_{\mathrm{TA},\mathbf{H}^{(L-1)\rightarrow L}}$ & 80.0 & 0.0 & 66.0 & 48.7 \\
    \rowcolor{highlight}
    $\mathrm{MergeVLA}_{\mathrm{TA},\mathbf{H}^{(L-2)\rightarrow L}}$ & 82.0 & 0.0 & 66.0 & 49.3\\
    \rowcolor{highlight}
    $\mathrm{MergeVLA}_{\mathrm{TIES},\mathbf{H}^{(L-1)\rightarrow L}}$ & \textbf{90.0} & 0.0 & \textbf{88.0} & 59.3 \\
    \rowcolor{highlight}
    $\mathrm{MergeVLA}_{\mathrm{TIES},\mathbf{H}^{(L-2)\rightarrow L}}$ & 88.0 & \textbf{38.0} & 86.0 & \textbf{70.7} \\
    \bottomrule
    \end{tabularx}
    \label{tab:robotwin}
\end{table}


\subsection{Results on LIBERO-Plus}
\label{sec:libero-plus}

Experiments on the LIBERO benchmark verify that MergeVLA achieves strong cross-task performance within known environments. 
To further examine the model’s \textit{generalization and robustness to unseen scenes}, we conduct additional evaluations on \textbf{LIBERO-Plus}, a variant of LIBERO that introduces controlled distribution shifts in both visual appearance and language descriptions. 
We use the same models trained on LIBERO and directly test them on the shifted LIBERO-Plus environments without any additional finetuning. 
This setting allows us to rigorously assess MergeVLA’s ability to maintain performance under out-of-distribution visual and linguistic conditions.

From Table \ref{tab:robustness}, we first observe that when using single-task finetuned checkpoints evaluated on their own tasks, MergeVLA exhibits stronger robustness under various perturbations than existing VLA models. This improvement stems from its architecture design, which preserves the pretrained VLM’s inherent robustness to visual and language perturbations. Secondly, in the model-merging setting, we find that applying different merging methods with MergeVLA maintains similar robustness even under cross-task evaluation. Notably, when using TA and TIES merging, the merged model even surpasses OpenVLA, $\mathrm{\pi_0}$, and VLA-Adapter in the single-task setting, demonstrating that our MergeVLA can effectively transfer and preserve robustness across tasks.

\subsection{Results on RoboTwin}

While evaluations on the LIBERO and LIBERO-Plus benchmarks demonstrate that MergeVLA attains high performance and robustness in cross-task settings, these tests are limited to a single embodiment. 
To examine the effectiveness of MergeVLA under \textit{cross-embodiment merging}, we selected \textbf{RoboTwin-2.0} as our evaluation platform, since it supports multiple dual-arm robots and enables a broad assessment of generalization across hardware.
Specifically, we designed two experimental settings: \textit{A:} Three dual-arm robots, Aloha-Agilex, ARX-X5, and Piper, each perform the same manipulation task \textsc{place container plate}. \textit{B:} The same three robots each perform a different task: \textsc{place container plate}, \textsc{handover block}, and \textsc{open microwave}, respectively.
For each combination of \{embodiment,task\}, we collected 50 demonstration trajectories for finetuning. 
We report the success rate of (i) single-\{task, embodiment\} fine-tuned models, and (ii) merged models obtained using TA and TIES strategies under the MergeVLA framework. 
The detailed results are presented in Table~\ref{tab:robotwin}, where each result is the success rate over 50 trials.
Unlike the experiments on LIBERO and LIBERO-Plus, merging across different embodiments in RoboTwin poses a greater challenge for the test-time task router.
We find that only keeping the final block $L$ as expert head is insufficient, as embodiment-specific differences in morphology and action space introduce stronger specialization and conflicts within the action heads.
For \textit{Setting A}, routing  $\mathbf{H}^{(L-1)\rightarrow L}$ preserves the performance of individually finetuned policies, indicating that the earlier merged blocks still capture transferable knowledge across different embodiments.
For \textit{Setting B}, routing $\mathbf{H}^{(L-2)\rightarrow L}$ and TIES merging are need to maintain comparable performance, especially for the \textsc{Handover Block} task, which requires coordinated dual-arm motion and thus induces stronger conflicts in the action space.
Overall, these findings highlight MergeVLA’s ability for cross-embodiment generalization.

\begin{figure}[t]
    \centering
    \includegraphics[width=0.45\textwidth]{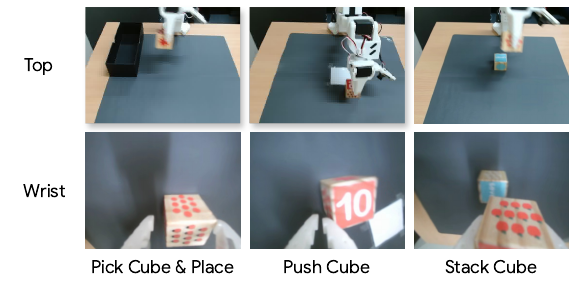}
    \caption{Setup of the real-world SO-101 arm experiments with three cube manipulation tasks.}
    \label{fig:env_real}
\end{figure}

\subsection{Real-World Experiments}
\label{sec:real-world}
\noindent\textbf{Tasks.}
As shown in Figure~\ref{fig:env_real}, we evaluate MergeVLA on three cube-based manipulation tasks using a real SO-101 robotic arm:

(i) \textsc{Pick \& Place}: the robot must grasp a cube and place it into a black box; success is recorded when the cube is stably placed inside the container.

(ii) \textsc{Push Cube}: the robot must push the cube into a designated white goal zone; success requires the cube to fully enter the region.

(iii) \textsc{Stack Cube}: the robot must pick up the red cube and place it on top of a blue cube; success is defined by a stable, non-slipping stacked configuration.

\noindent\textbf{Data Collection.}
We collect demonstrations using the SO-101 arm under a leader–follower teleoperation setup, with two RGB camera views: a fixed top-down camera and a wrist-mounted camera.  
For each task, we collect 50 demonstrations at a frequency of 20 Hz, with randomized cube starting positions. For the \textsc{Pick \& Place} and \textsc{Push Cube} tasks, only the red cube is used during data collection.
Each demonstration includes synchronized RGB observations, 6 DoF joint actions, and task instructions.
We train MergeVLA for 30k steps per task.

\noindent\textbf{Evaluation Protocol.}
For each model to be evaluated, we perform 20 rollouts per task, with randomized cube initial positions in every rollout. For \textsc{Pick \& Place} and \textsc{Push Cube} tasks, we use cubes with randomly different colors that are unseen in the training data, providing a visual shift evaluation.
Success is determined according to the task-specific criteria defined above.
We report the success rate as the percentage of successful trials out of the 20 rollouts.

\noindent\textbf{Results.}
In Table~\ref{tab:real}, we present both fine-tuning and model-merging results of MergeVLA on the real SO-101 robotic arm.
For fine-tuning, MergeVLA achieves high success rates across all three cube-manipulation tasks. Notably, in \textsc{Pick \& Place} and \textsc{Push Cube}, the robot is required to operate on cubes whose colors differ from those seen during training. MergeVLA remains robust under this distribution shift, reliably detecting the target object and executing the required manipulation, which highlights its strong visual out-of-distribution generalization in real-world settings.

For model merging, we evaluate MergeVLA using TA and TIES as the merging strategies. TIES-based merging delivers the best overall performance, often matching the results of the corresponding single-task models. This demonstrates that MergeVLA preserves cross-task merging ability even when deployed on physical hardware, and is able to reuse skill components without degradation, an encouraging indication of its practicality for multi-skill real robot systems.

\begin{table}[t]
    \Large
    \centering
    \caption{Real-world SO-101 robot performance, reported as success rates (\%) over 20 rollouts per task.}
    \label{tab:real}
    \resizebox{\columnwidth}{!}{
    \begin{tabular}{lcccc}
        \toprule
        \textbf{Method} & \textbf{Pick \& Place} & \textbf{Push Cube} & \textbf{Stack Cube} & \textbf{Avg.} \\
        \midrule
        \rowcolor{highlightgray}
        Single-task finetune & 90.0 & 85.0 & 95.0 & 90.0\\
        \rowcolor{highlight}
        $\mathrm{MergeVLA}_{\mathrm{TA}}$ & 70.0 & 70.0 & 60.0 & 66.7 \\
        \rowcolor{highlight}
        $\mathrm{MergeVLA}_{\mathrm{TIES}}$ & \textbf{90.0} & \textbf{90.0} & \textbf{90.0} & \textbf{90.0} \\

        \bottomrule
    \end{tabular}
    }
\end{table}

\subsection{Ablation Study}

\begin{figure}[t]
        \centering
        \begin{subfigure}[t]{.41\linewidth}
            \centering
            \includegraphics[width=1\linewidth]{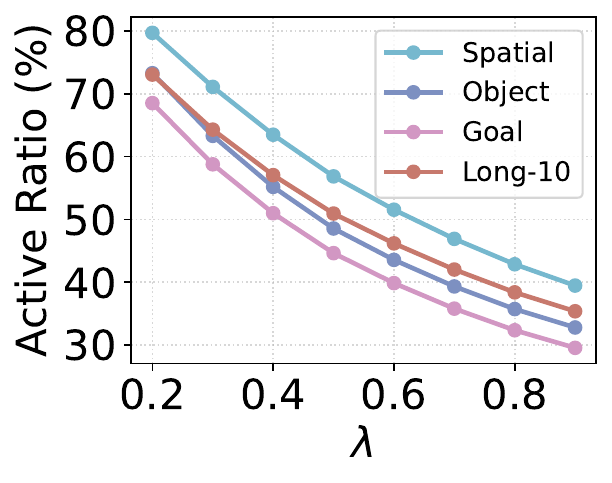}
        \end{subfigure}
        \begin{subfigure}[t]{.57\linewidth}
            \centering
            \includegraphics[width=1\linewidth]{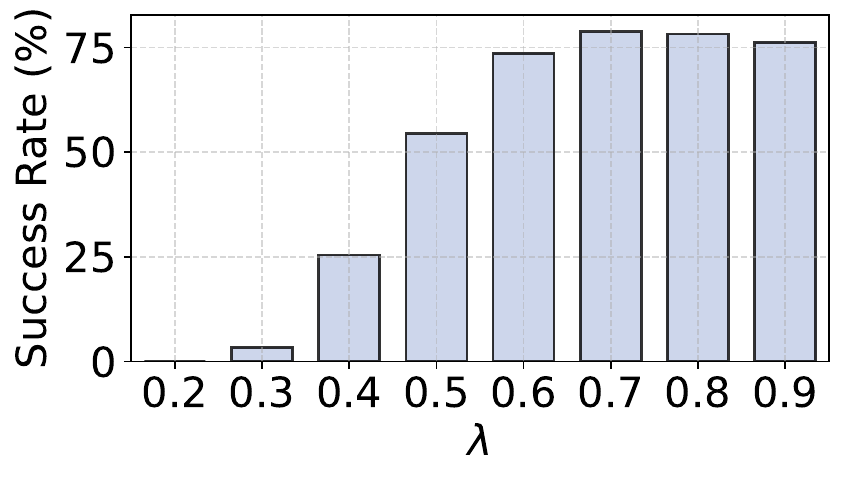}
        \end{subfigure}
        \caption{The ablation study and analysis of $\lambda$. (a) The mask active ratio across different $\lambda$ of LIBERO. (b) The success rate across different $\lambda$ of LIBERO-Long.}
        \label{fig:mask ratio}
\end{figure}

\noindent \textbf{Impact of Mask Ratio.} We analyze the effect of the task mask under different $\lambda$, which controls the active ratio of the mask. We vary $\lambda$ from 0.2 to 0.9 and obtain the merged task vector using Task Arithmetic (TA). Figure~\ref{fig:mask ratio} (a) shows the active ratio of the mask across four tasks. As $\lambda$ increases, fewer parameters remain active, with the LIBERO-Spatial task consistently showing the highest activation ratio, indicating its dominant weight contribution. We further evaluate the impact of $\lambda$ on the model’s performance in the LIBERO-Long task, as shown in Figure~\ref{fig:mask ratio} (b). When $\lambda$ is small (\textit{e.g.,} 0.2), the mask activates too many parameters, leading to severe task interference and even complete failure. In contrast, when $\lambda$ lies between 0.6 and 0.9, the success rate exceeds 70\%. These results suggest that moderate sparsity enables an effective balance between task-specific and merged task vectors. When the parameter differences across tasks are large, relying more on the pretrained model yields better results, whereas emphasizing the merged task vector becomes effective only when the task-specific weights dominate.

\noindent \textbf{Impact of the Subspace Used for Routing.} To validate our task router design, we experiment on all four LIBERO task suites using three configurations: (1) using $\mathbf{K}$ projections, (2) using $\mathbf{V}$ projections, and (3) using $\mathbf{K}$ and $\mathbf{V}$ projections jointly, while fixing $\lambda=0.6$ for the task-specific mask and adopting TA for VLM merging. The results are summarized in Table \ref{tab:libero_routing_key}. We observe that for Spatial and Long tasks, the choice of projection combination has little effect on performance. Yet, for Object and Goal, using only $\mathbf{K}$ or combining $\mathbf{K}$ and $\mathbf{V}$ leads to a dramatic drop in success rate, and even complete failure in some cases. By inspecting the router’s task selection, we find that when using $\mathbf{K}$, the router tends to misassign tasks. From the perspective of attention interaction, the value projection captures the actual behavioral semantics retrieved by the query, whereas the key projection primarily defines the similarity structure of the query. Consequently, value-based alignment provides a more reliable indicator of task identity.

\begin{table}[t]
    \small
    \centering
    \caption{Ablation results of MergeVLA with different subspaces used for routing on LIBERO.}
    \label{tab:libero_routing_key}
    \begin{tabular}{lccccc}
        \toprule
        \textbf{Method} & \textbf{Spatial} & \textbf{Object} & \textbf{Goal} & \textbf{Long} & \textbf{Avg.} \\
        \midrule
        Only $\mathbf{K}$& \textbf{98.0} & 0.0& 39.6& \textbf{76.6} & 53.6\\
        $\mathbf{K}\&\mathbf{V}$ & \textbf{98.0} & 0.0& \textbf{85.8} & \textbf{76.6} & 65.1\\
        \rowcolor{highlight}
        Only $\mathbf{V}$ &  \textbf{98.0} &  \textbf{98.8} &  85.4&  \textbf{76.6} &  \textbf{89.7}\\
        \bottomrule
    \end{tabular}
\end{table}

\section{Conclusion}
The lack of cross-skill capability in existing VLA models remains a critical yet unexplored challenge. We show that current model merging techniques cannot be directly applied to VLAs due to destructive LoRA parameter interference and architectural incompatibility of action experts. We propose MergeVLA, a framework for merging VLA models that is compatible with mainstream merging methods toward an embodied generalist. Experiments across three benchmarks demonstrate MergeVLA’s strong multi-task ability and robustness under distribution shifts, with transferability across embodiments. Future work will explore this promising direction further, including whether larger VLM backbones remain compatible with our framework and whether pretraining on diverse robot datasets can further enhance merging effectiveness.

\section{Acknowledgments}
This work was partially supported by ARC DE240100105, DP240101814, DP230101196, BA24006, and ARC Industrial Transformation Research Hubs IH230100013.

\newpage
{
    \small
    \bibliographystyle{ieeenat_fullname}
    \bibliography{main}

@String(CVPR= {IEEE Conf. Comput. Vis. Pattern Recog.})

@String(ECCV= {Eur. Conf. Comput. Vis.})

@String(ICLR = {Int. Conf. Learn. Represent.})

@String(CVPR  = {CVPR})

@String(ECCV  = {ECCV})

@String(ICLR  = {ICLR})

@inproceedings{OpenVLA,
  author       = {Moo Jin Kim and
                  Karl Pertsch and
                  Siddharth Karamcheti and
                  Ted Xiao and
                  Ashwin Balakrishna and
                  Suraj Nair and
                  Rafael Rafailov and
                  Ethan Paul Foster and
                  Pannag R. Sanketi and
                  Quan Vuong and
                  Thomas Kollar and
                  Benjamin Burchfiel and
                  Russ Tedrake and
                  Dorsa Sadigh and
                  Sergey Levine and
                  Percy Liang and
                  Chelsea Finn},
  editor       = {Pulkit Agrawal and
                  Oliver Kroemer and
                  Wolfram Burgard},
  title        = {OpenVLA: An Open-Source Vision-Language-Action Model},
  booktitle    = {Conference on Robot Learning, 6-9 November 2024, Munich, Germany},
  series       = {Proceedings of Machine Learning Research},
  volume       = {270},
  pages        = {2679--2713},
  publisher    = {{PMLR}},
  year         = {2024},
  url          = {https://proceedings.mlr.press/v270/kim25c.html},
  bibsource    = {dblp computer science bibliography, https://dblp.org}
}

@article{pi0,
  author       = {Kevin Black and
                  Noah Brown and
                  Danny Driess and
                  Adnan Esmail and
                  Michael Equi and
                  Chelsea Finn and
                  Niccolo Fusai and
                  Lachy Groom and
                  Karol Hausman and
                  Brian Ichter and
                  Szymon Jakubczak and
                  Tim Jones and
                  Liyiming Ke and
                  Sergey Levine and
                  Adrian Li{-}Bell and
                  Mohith Mothukuri and
                  Suraj Nair and
                  Karl Pertsch and
                  Lucy Xiaoyang Shi and
                  James Tanner and
                  Quan Vuong and
                  Anna Walling and
                  Haohuan Wang and
                  Ury Zhilinsky},
  title        = {{\(\pi\)}\({}_{\mbox{0}}\): {A} Vision-Language-Action Flow Model
                  for General Robot Control},
  journal      = {CoRR},
  volume       = {abs/2410.24164},
  year         = {2024},
  url          = {https://doi.org/10.48550/arXiv.2410.24164},
  doi          = {10.48550/ARXIV.2410.24164},
  eprinttype    = {arXiv},
  eprint       = {2410.24164},
  bibsource    = {dblp computer science bibliography, https://dblp.org}
}

@article{pi0.5,
  author       = {Physical Intelligence and
                  Kevin Black and
                  Noah Brown and
                  James Darpinian and
                  Karan Dhabalia and
                  Danny Driess and
                  Adnan Esmail and
                  Michael Equi and
                  Chelsea Finn and
                  Niccolo Fusai and
                  Manuel Y. Galliker and
                  Dibya Ghosh and
                  Lachy Groom and
                  Karol Hausman and
                  Brian Ichter and
                  Szymon Jakubczak and
                  Tim Jones and
                  Liyiming Ke and
                  Devin LeBlanc and
                  Sergey Levine and
                  Adrian Li{-}Bell and
                  Mohith Mothukuri and
                  Suraj Nair and
                  Karl Pertsch and
                  Allen Z. Ren and
                  Lucy Xiaoyang Shi and
                  Laura Smith and
                  Jost Tobias Springenberg and
                  Kyle Stachowicz and
                  James Tanner and
                  Quan Vuong and
                  Homer Walke and
                  Anna Walling and
                  Haohuan Wang and
                  Lili Yu and
                  Ury Zhilinsky},
  title        = {{\(\pi\)}\({}_{\mbox{0.5}}\): a Vision-Language-Action Model with
                  Open-World Generalization},
  journal      = {CoRR},
  volume       = {abs/2504.16054},
  year         = {2025},
  url          = {https://doi.org/10.48550/arXiv.2504.16054},
  doi          = {10.48550/ARXIV.2504.16054},
  eprinttype    = {arXiv},
  eprint       = {2504.16054},
  bibsource    = {dblp computer science bibliography, https://dblp.org}
}

@inproceedings{libero,
  author       = {Bo Liu and
                  Yifeng Zhu and
                  Chongkai Gao and
                  Yihao Feng and
                  Qiang Liu and
                  Yuke Zhu and
                  Peter Stone},
  editor       = {Alice Oh and
                  Tristan Naumann and
                  Amir Globerson and
                  Kate Saenko and
                  Moritz Hardt and
                  Sergey Levine},
  title        = {{LIBERO:} Benchmarking Knowledge Transfer for Lifelong Robot Learning},
  booktitle    = {Advances in Neural Information Processing Systems 36: Annual Conference
                  on Neural Information Processing Systems 2023, NeurIPS 2023, New Orleans,
                  LA, USA, December 10 - 16, 2023},
  year         = {2023},
  url          = {http://papers.nips.cc/paper\_files/paper/2023/hash/8c3c666820ea055a77726d66fc7d447f-Abstract-Datasets\_and\_Benchmarks.html},
  bibsource    = {dblp computer science bibliography, https://dblp.org}
}

@misc{liberoplus,
      title={LIBERO-Plus: In-depth Robustness Analysis of Vision-Language-Action Models}, 
      author={Senyu Fei and Siyin Wang and Junhao Shi and Zihao Dai and Jikun Cai and Pengfang Qian and Li Ji and Xinzhe He and Shiduo Zhang and Zhaoye Fei and Jinlan Fu and Jingjing Gong and Xipeng Qiu},
      year={2025},
      eprint={2510.13626},
      archivePrefix={arXiv},
      primaryClass={cs.RO},
      url={https://arxiv.org/abs/2510.13626}, 
}

@article{robotwin,
  author       = {Tianxing Chen and
                  Zanxin Chen and
                  Baijun Chen and
                  Zijian Cai and
                  Yibin Liu and
                  Qiwei Liang and
                  Zixuan Li and
                  Xianliang Lin and
                  Yiheng Ge and
                  Zhenyu Gu and
                  Weiliang Deng and
                  Yubin Guo and
                  Tian Nian and
                  Xuanbing Xie and
                  Qiangyu Chen and
                  Kailun Su and
                  Tianling Xu and
                  Guodong Liu and
                  Mengkang Hu and
                  Huan{-}ang Gao and
                  Kaixuan Wang and
                  Zhixuan Liang and
                  Yusen Qin and
                  Xiaokang Yang and
                  Ping Luo and
                  Yao Mu},
  title        = {RoboTwin 2.0: {A} Scalable Data Generator and Benchmark with Strong
                  Domain Randomization for Robust Bimanual Robotic Manipulation},
  journal      = {CoRR},
  volume       = {abs/2506.18088},
  year         = {2025},
  url          = {https://doi.org/10.48550/arXiv.2506.18088},
  doi          = {10.48550/ARXIV.2506.18088},
  eprinttype    = {arXiv},
  eprint       = {2506.18088},
  bibsource    = {dblp computer science bibliography, https://dblp.org}
}

@inproceedings{rt2,
  author       = {Brianna Zitkovich and
                  Tianhe Yu and
                  Sichun Xu and
                  Peng Xu and
                  Ted Xiao and
                  Fei Xia and
                  Jialin Wu and
                  Paul Wohlhart and
                  Stefan Welker and
                  Ayzaan Wahid and
                  Quan Vuong and
                  Vincent Vanhoucke and
                  Huong T. Tran and
                  Radu Soricut and
                  Anikait Singh and
                  Jaspiar Singh and
                  Pierre Sermanet and
                  Pannag R. Sanketi and
                  Grecia Salazar and
                  Michael S. Ryoo and
                  Krista Reymann and
                  Kanishka Rao and
                  Karl Pertsch and
                  Igor Mordatch and
                  Henryk Michalewski and
                  Yao Lu and
                  Sergey Levine and
                  Lisa Lee and
                  Tsang{-}Wei Edward Lee and
                  Isabel Leal and
                  Yuheng Kuang and
                  Dmitry Kalashnikov and
                  Ryan Julian and
                  Nikhil J. Joshi and
                  Alex Irpan and
                  Brian Ichter and
                  Jasmine Hsu and
                  Alexander Herzog and
                  Karol Hausman and
                  Keerthana Gopalakrishnan and
                  Chuyuan Fu and
                  Pete Florence and
                  Chelsea Finn and
                  Kumar Avinava Dubey and
                  Danny Driess and
                  Tianli Ding and
                  Krzysztof Marcin Choromanski and
                  Xi Chen and
                  Yevgen Chebotar and
                  Justice Carbajal and
                  Noah Brown and
                  Anthony Brohan and
                  Montserrat Gonzalez Arenas and
                  Kehang Han},
  editor       = {Jie Tan and
                  Marc Toussaint and
                  Kourosh Darvish},
  title        = {{RT-2:} Vision-Language-Action Models Transfer Web Knowledge to Robotic
                  Control},
  booktitle    = {Conference on Robot Learning, CoRL 2023, 6-9 November 2023, Atlanta,
                  GA, {USA}},
  series       = {Proceedings of Machine Learning Research},
  volume       = {229},
  pages        = {2165--2183},
  publisher    = {{PMLR}},
  year         = {2023},
  url          = {https://proceedings.mlr.press/v229/zitkovich23a.html},
  bibsource    = {dblp computer science bibliography, https://dblp.org}
}

@article{vla-adapter,
  author       = {Yihao Wang and
                  Pengxiang Ding and
                  Lingxiao Li and
                  Can Cui and
                  Zirui Ge and
                  Xinyang Tong and
                  Wenxuan Song and
                  Han Zhao and
                  Wei Zhao and
                  Pengxu Hou and
                  Siteng Huang and
                  Yifan Tang and
                  Wenhui Wang and
                  Ru Zhang and
                  Jianyi Liu and
                  Donglin Wang},
  title        = {VLA-Adapter: An Effective Paradigm for Tiny-Scale Vision-Language-Action
                  Model},
  journal      = {CoRR},
  volume       = {abs/2509.09372},
  year         = {2025},
  url          = {https://doi.org/10.48550/arXiv.2509.09372},
  doi          = {10.48550/ARXIV.2509.09372},
  eprinttype    = {arXiv},
  eprint       = {2509.09372},
  bibsource    = {dblp computer science bibliography, https://dblp.org}
}

@inproceedings{bridgedatav2,
  author       = {Homer Rich Walke and
                  Kevin Black and
                  Tony Z. Zhao and
                  Quan Vuong and
                  Chongyi Zheng and
                  Philippe Hansen{-}Estruch and
                  Andre Wang He and
                  Vivek Myers and
                  Moo Jin Kim and
                  Max Du and
                  Abraham Lee and
                  Kuan Fang and
                  Chelsea Finn and
                  Sergey Levine},
  editor       = {Jie Tan and
                  Marc Toussaint and
                  Kourosh Darvish},
  title        = {BridgeData {V2:} {A} Dataset for Robot Learning at Scale},
  booktitle    = {Conference on Robot Learning, CoRL 2023, 6-9 November 2023, Atlanta,
                  GA, {USA}},
  series       = {Proceedings of Machine Learning Research},
  volume       = {229},
  pages        = {1723--1736},
  publisher    = {{PMLR}},
  year         = {2023},
  url          = {https://proceedings.mlr.press/v229/walke23a.html},
  bibsource    = {dblp computer science bibliography, https://dblp.org}
}

@inproceedings{bridgedata,
  author       = {Frederik Ebert and
                  Yanlai Yang and
                  Karl Schmeckpeper and
                  Bernadette Bucher and
                  Georgios Georgakis and
                  Kostas Daniilidis and
                  Chelsea Finn and
                  Sergey Levine},
  editor       = {Kris Hauser and
                  Dylan A. Shell and
                  Shoudong Huang},
  title        = {Bridge Data: Boosting Generalization of Robotic Skills with Cross-Domain
                  Datasets},
  booktitle    = {Robotics: Science and Systems XVIII, New York City, NY, USA, June
                  27 - July 1, 2022},
  year         = {2022},
  url          = {https://doi.org/10.15607/RSS.2022.XVIII.063},
  doi          = {10.15607/RSS.2022.XVIII.063},
  bibsource    = {dblp computer science bibliography, https://dblp.org}
}

@inproceedings{OpenXEmbodiment,
  author       = {Abby O'Neill and
                  Abdul Rehman and
                  Abhiram Maddukuri and
                  Abhishek Gupta and
                  Abhishek Padalkar and
                  Abraham Lee and
                  Acorn Pooley and
                  Agrim Gupta and
                  Ajay Mandlekar and
                  Ajinkya Jain and
                  Albert Tung and
                  Alex Bewley and
                  Alexander Herzog and
                  Alex Irpan and
                  Alexander Khazatsky and
                  Anant Rai and
                  Anchit Gupta and
                  Andrew E. Wang and
                  Anikait Singh and
                  Animesh Garg and
                  Aniruddha Kembhavi and
                  Annie Xie and
                  Anthony Brohan and
                  Antonin Raffin and
                  Archit Sharma and
                  Arefeh Yavary and
                  Arhan Jain and
                  Ashwin Balakrishna and
                  Ayzaan Wahid and
                  Ben Burgess{-}Limerick and
                  Beomjoon Kim and
                  Bernhard Sch{\"{o}}lkopf and
                  Blake Wulfe and
                  Brian Ichter and
                  Cewu Lu and
                  Charles Xu and
                  Charlotte Le and
                  Chelsea Finn and
                  Chen Wang and
                  Chenfeng Xu and
                  Cheng Chi and
                  Chenguang Huang and
                  Christine Chan and
                  Christopher Agia and
                  Chuer Pan and
                  Chuyuan Fu and
                  Coline Devin and
                  Danfei Xu and
                  Daniel Morton and
                  Danny Driess and
                  Daphne Chen and
                  Deepak Pathak and
                  Dhruv Shah and
                  Dieter B{\"{u}}chler and
                  Dinesh Jayaraman and
                  Dmitry Kalashnikov and
                  Dorsa Sadigh and
                  Edward Johns and
                  Ethan Paul Foster and
                  Fangchen Liu and
                  Federico Ceola and
                  Fei Xia and
                  Feiyu Zhao and
                  Freek Stulp and
                  Gaoyue Zhou and
                  Gaurav S. Sukhatme and
                  Gautam Salhotra and
                  Ge Yan and
                  Gilbert Feng and
                  Giulio Schiavi and
                  Glen Berseth and
                  Gregory Kahn and
                  Guanzhi Wang and
                  Hao Su and
                  Haoshu Fang and
                  Haochen Shi and
                  Henghui Bao and
                  Heni Ben Amor and
                  Henrik I. Christensen and
                  Hiroki Furuta and
                  Homer Walke and
                  Hongjie Fang and
                  Huy Ha and
                  Igor Mordatch and
                  Ilija Radosavovic and
                  Isabel Leal and
                  Jacky Liang and
                  Jad Abou{-}Chakra and
                  Jaehyung Kim and
                  Jaimyn Drake and
                  Jan Peters and
                  Jan Schneider and
                  Jasmine Hsu and
                  Jeannette Bohg and
                  Jeffrey T. Bingham and
                  Jeffrey Wu and
                  Jensen Gao and
                  Jiaheng Hu and
                  Jiajun Wu and
                  Jialin Wu and
                  Jiankai Sun and
                  Jianlan Luo and
                  Jiayuan Gu and
                  Jie Tan and
                  Jihoon Oh and
                  Jimmy Wu and
                  Jingpei Lu and
                  Jingyun Yang and
                  Jitendra Malik and
                  Jo{\~{a}}o Silv{\'{e}}rio and
                  Joey Hejna and
                  Jonathan Booher and
                  Jonathan Tompson and
                  Jonathan Yang and
                  Jordi Salvador and
                  Joseph J. Lim and
                  Junhyek Han and
                  Kaiyuan Wang and
                  Kanishka Rao and
                  Karl Pertsch and
                  Karol Hausman and
                  Keegan Go and
                  Keerthana Gopalakrishnan and
                  Ken Goldberg and
                  Kendra Byrne and
                  Kenneth Oslund and
                  Kento Kawaharazuka and
                  Kevin Black and
                  Kevin Lin and
                  Kevin Zhang and
                  Kiana Ehsani and
                  Kiran Lekkala and
                  Kirsty Ellis and
                  Krishan Rana and
                  Krishnan Srinivasan and
                  Kuan Fang and
                  Kunal Pratap Singh and
                  Kuo{-}Hao Zeng and
                  Kyle Hatch and
                  Kyle Hsu and
                  Laurent Itti and
                  Lawrence Yunliang Chen and
                  Lerrel Pinto and
                  Li Fei{-}Fei and
                  Liam Tan and
                  Linxi Jim Fan and
                  Lionel Ott and
                  Lisa Lee and
                  Luca Weihs and
                  Magnum Chen and
                  Marion Lepert and
                  Marius Memmel and
                  Masayoshi Tomizuka and
                  Masha Itkina and
                  Mateo Guaman Castro and
                  Max Spero and
                  Maximilian Du and
                  Michael Ahn and
                  Michael C. Yip and
                  Mingtong Zhang and
                  Mingyu Ding and
                  Minho Heo and
                  Mohan Kumar Srirama and
                  Mohit Sharma and
                  Moo Jin Kim and
                  Naoaki Kanazawa and
                  Nicklas Hansen and
                  Nicolas Heess and
                  Nikhil J. Joshi and
                  Niko S{\"{u}}nderhauf and
                  Ning Liu and
                  Norman Di Palo and
                  Nur Muhammad (Mahi) Shafiullah and
                  Oier Mees and
                  Oliver Kroemer and
                  Osbert Bastani and
                  Pannag R. Sanketi and
                  Patrick Tree Miller and
                  Patrick Yin and
                  Paul Wohlhart and
                  Peng Xu and
                  Peter David Fagan and
                  Peter Mitrano and
                  Pierre Sermanet and
                  Pieter Abbeel and
                  Priya Sundaresan and
                  Qiuyu Chen and
                  Quan Vuong and
                  Rafael Rafailov and
                  Ran Tian and
                  Ria Doshi and
                  Roberto Mart{\'{\i}}n{-}Mart{\'{\i}}n and
                  Rohan Baijal and
                  Rosario Scalise and
                  Rose Hendrix and
                  Roy Lin and
                  Runjia Qian and
                  Ruohan Zhang and
                  Russell Mendonca and
                  Rutav Shah and
                  Ryan Hoque and
                  Ryan Julian and
                  Samuel Bustamante{-}Gomez and
                  Sean Kirmani and
                  Sergey Levine and
                  Shan Lin and
                  Sherry Moore and
                  Shikhar Bahl and
                  Shivin Dass and
                  Shubham D. Sonawani and
                  Shuran Song and
                  Sichun Xu and
                  Siddhant Haldar and
                  Siddharth Karamcheti and
                  Simeon Adebola and
                  Simon Guist and
                  Soroush Nasiriany and
                  Stefan Schaal and
                  Stefan Welker and
                  Stephen Tian and
                  Subramanian Ramamoorthy and
                  Sudeep Dasari and
                  Suneel Belkhale and
                  Sungjae Park and
                  Suraj Nair and
                  Suvir Mirchandani and
                  Takayuki Osa and
                  Tanmay Gupta and
                  Tatsuya Harada and
                  Tatsuya Matsushima and
                  Ted Xiao and
                  Thomas Kollar and
                  Tianhe Yu and
                  Tianli Ding and
                  Todor Davchev and
                  Tony Z. Zhao and
                  Travis Armstrong and
                  Trevor Darrell and
                  Trinity Chung and
                  Vidhi Jain and
                  Vincent Vanhoucke and
                  Wei Zhan and
                  Wenxuan Zhou and
                  Wolfram Burgard and
                  Xi Chen and
                  Xiaolong Wang and
                  Xinghao Zhu and
                  Xinyang Geng and
                  Xiyuan Liu and
                  Liangwei Xu and
                  Xuanlin Li and
                  Yao Lu and
                  Yecheng Jason Ma and
                  Yejin Kim and
                  Yevgen Chebotar and
                  Yifan Zhou and
                  Yifeng Zhu and
                  Yilin Wu and
                  Ying Xu and
                  Yixuan Wang and
                  Yonatan Bisk and
                  Yoonyoung Cho and
                  Youngwoon Lee and
                  Yuchen Cui and
                  Yue Cao and
                  Yueh{-}Hua Wu and
                  Yujin Tang and
                  Yuke Zhu and
                  Yunchu Zhang and
                  Yunfan Jiang and
                  Yunshuang Li and
                  Yunzhu Li and
                  Yusuke Iwasawa and
                  Yutaka Matsuo and
                  Zehan Ma and
                  Zhuo Xu and
                  Zichen Jeff Cui and
                  Zichen Zhang and
                  Zipeng Lin},
  title        = {Open X-Embodiment: Robotic Learning Datasets and {RT-X} Models : Open
                  X-Embodiment Collaboration},
  booktitle    = {{IEEE} International Conference on Robotics and Automation, {ICRA}
                  2024, Yokohama, Japan, May 13-17, 2024},
  pages        = {6892--6903},
  publisher    = {{IEEE}},
  year         = {2024},
  url          = {https://doi.org/10.1109/ICRA57147.2024.10611477},
  doi          = {10.1109/ICRA57147.2024.10611477},
  bibsource    = {dblp computer science bibliography, https://dblp.org}
}

@article{calvin,
  author       = {Oier Mees and
                  Luk{\'{a}}s Hermann and
                  Erick Rosete{-}Beas and
                  Wolfram Burgard},
  title        = {{CALVIN:} {A} Benchmark for Language-Conditioned Policy Learning for
                  Long-Horizon Robot Manipulation Tasks},
  journal      = {{IEEE} Robotics Autom. Lett.},
  volume       = {7},
  number       = {3},
  pages        = {7327--7334},
  year         = {2022},
  url          = {https://doi.org/10.1109/LRA.2022.3180108},
  doi          = {10.1109/LRA.2022.3180108},
  bibsource    = {dblp computer science bibliography, https://dblp.org}
}

@article{maniskill,
  author       = {Stone Tao and
                  Fanbo Xiang and
                  Arth Shukla and
                  Yuzhe Qin and
                  Xander Hinrichsen and
                  Xiaodi Yuan and
                  Chen Bao and
                  Xinsong Lin and
                  Yulin Liu and
                  Tse{-}kai Chan and
                  Yuan Gao and
                  Xuanlin Li and
                  Tongzhou Mu and
                  Nan Xiao and
                  Arnav Gurha and
                  Zhiao Huang and
                  Roberto Calandra and
                  Rui Chen and
                  Shan Luo and
                  Hao Su},
  title        = {ManiSkill3: {GPU} Parallelized Robotics Simulation and Rendering for
                  Generalizable Embodied {AI}},
  journal      = {CoRR},
  volume       = {abs/2410.00425},
  year         = {2024},
  url          = {https://doi.org/10.48550/arXiv.2410.00425},
  doi          = {10.48550/ARXIV.2410.00425},
  eprinttype    = {arXiv},
  eprint       = {2410.00425},
  bibsource    = {dblp computer science bibliography, https://dblp.org}
}

@inproceedings{prismatic,
  author       = {Siddharth Karamcheti and
                  Suraj Nair and
                  Ashwin Balakrishna and
                  Percy Liang and
                  Thomas Kollar and
                  Dorsa Sadigh},
  title        = {Prismatic VLMs: Investigating the Design Space of Visually-Conditioned
                  Language Models},
  booktitle    = {Forty-first International Conference on Machine Learning, {ICML} 2024,
                  Vienna, Austria, July 21-27, 2024},
  publisher    = {OpenReview.net},
  year         = {2024},
  url          = {https://openreview.net/forum?id=6FXtu8clyp},
  bibsource    = {dblp computer science bibliography, https://dblp.org}
}

@inproceedings{rt1,
  author       = {Anthony Brohan and
                  Noah Brown and
                  Justice Carbajal and
                  Yevgen Chebotar and
                  Joseph Dabis and
                  Chelsea Finn and
                  Keerthana Gopalakrishnan and
                  Karol Hausman and
                  Alexander Herzog and
                  Jasmine Hsu and
                  Julian Ibarz and
                  Brian Ichter and
                  Alex Irpan and
                  Tomas Jackson and
                  Sally Jesmonth and
                  Nikhil J. Joshi and
                  Ryan Julian and
                  Dmitry Kalashnikov and
                  Yuheng Kuang and
                  Isabel Leal and
                  Kuang{-}Huei Lee and
                  Sergey Levine and
                  Yao Lu and
                  Utsav Malla and
                  Deeksha Manjunath and
                  Igor Mordatch and
                  Ofir Nachum and
                  Carolina Parada and
                  Jodilyn Peralta and
                  Emily Perez and
                  Karl Pertsch and
                  Jornell Quiambao and
                  Kanishka Rao and
                  Michael S. Ryoo and
                  Grecia Salazar and
                  Pannag R. Sanketi and
                  Kevin Sayed and
                  Jaspiar Singh and
                  Sumedh Sontakke and
                  Austin Stone and
                  Clayton Tan and
                  Huong T. Tran and
                  Vincent Vanhoucke and
                  Steve Vega and
                  Quan Vuong and
                  Fei Xia and
                  Ted Xiao and
                  Peng Xu and
                  Sichun Xu and
                  Tianhe Yu and
                  Brianna Zitkovich},
  editor       = {Kostas E. Bekris and
                  Kris Hauser and
                  Sylvia L. Herbert and
                  Jingjin Yu},
  title        = {{RT-1:} Robotics Transformer for Real-World Control at Scale},
  booktitle    = {Robotics: Science and Systems XIX, Daegu, Republic of Korea, July
                  10-14, 2023},
  year         = {2023},
  url          = {https://doi.org/10.15607/RSS.2023.XIX.025},
  doi          = {10.15607/RSS.2023.XIX.025},
  bibsource    = {dblp computer science bibliography, https://dblp.org}
}

@article{openhelix,
  author       = {Can Cui and
                  Pengxiang Ding and
                  Wenxuan Song and
                  Shuanghao Bai and
                  Xinyang Tong and
                  Zirui Ge and
                  Runze Suo and
                  Wanqi Zhou and
                  Yang Liu and
                  Bofang Jia and
                  Han Zhao and
                  Siteng Huang and
                  Donglin Wang},
  title        = {OpenHelix: {A} Short Survey, Empirical Analysis, and Open-Source Dual-System
                  {VLA} Model for Robotic Manipulation},
  journal      = {CoRR},
  volume       = {abs/2505.03912},
  year         = {2025},
  url          = {https://doi.org/10.48550/arXiv.2505.03912},
  doi          = {10.48550/ARXIV.2505.03912},
  eprinttype    = {arXiv},
  eprint       = {2505.03912},
  bibsource    = {dblp computer science bibliography, https://dblp.org}
}

@article{dual1,
  author       = {Qingwen Bu and
                  Hongyang Li and
                  Li Chen and
                  Jisong Cai and
                  Jia Zeng and
                  Heming Cui and
                  Maoqing Yao and
                  Yu Qiao},
  title        = {Towards Synergistic, Generalized, and Efficient Dual-System for Robotic
                  Manipulation},
  journal      = {CoRR},
  volume       = {abs/2410.08001},
  year         = {2024},
  url          = {https://doi.org/10.48550/arXiv.2410.08001},
  doi          = {10.48550/ARXIV.2410.08001},
  eprinttype    = {arXiv},
  eprint       = {2410.08001},
  bibsource    = {dblp computer science bibliography, https://dblp.org}
}

@inproceedings{hirt,
  author       = {Jianke Zhang and
                  Yanjiang Guo and
                  Xiaoyu Chen and
                  Yen{-}Jen Wang and
                  Yucheng Hu and
                  Chengming Shi and
                  Jianyu Chen},
  editor       = {Pulkit Agrawal and
                  Oliver Kroemer and
                  Wolfram Burgard},
  title        = {HiRT: Enhancing Robotic Control with Hierarchical Robot Transformers},
  booktitle    = {Conference on Robot Learning, 6-9 November 2024, Munich, Germany},
  series       = {Proceedings of Machine Learning Research},
  volume       = {270},
  pages        = {933--946},
  publisher    = {{PMLR}},
  year         = {2024},
  url          = {https://proceedings.mlr.press/v270/zhang25b.html},
  bibsource    = {dblp computer science bibliography, https://dblp.org}
}

@article{mtopt,
  author       = {Dmitry Kalashnikov and
                  Jacob Varley and
                  Yevgen Chebotar and
                  Benjamin Swanson and
                  Rico Jonschkowski and
                  Chelsea Finn and
                  Sergey Levine and
                  Karol Hausman},
  title        = {MT-Opt: Continuous Multi-Task Robotic Reinforcement Learning at Scale},
  journal      = {CoRR},
  volume       = {abs/2104.08212},
  year         = {2021},
  url          = {https://arxiv.org/abs/2104.08212},
  eprinttype    = {arXiv},
  eprint       = {2104.08212},
  bibsource    = {dblp computer science bibliography, https://dblp.org}
}

@article{qtopt,
  author       = {Dmitry Kalashnikov and
                  Alex Irpan and
                  Peter Pastor and
                  Julian Ibarz and
                  Alexander Herzog and
                  Eric Jang and
                  Deirdre Quillen and
                  Ethan Holly and
                  Mrinal Kalakrishnan and
                  Vincent Vanhoucke and
                  Sergey Levine},
  title        = {QT-Opt: Scalable Deep Reinforcement Learning for Vision-Based Robotic
                  Manipulation},
  journal      = {CoRR},
  volume       = {abs/1806.10293},
  year         = {2018},
  url          = {http://arxiv.org/abs/1806.10293},
  eprinttype    = {arXiv},
  eprint       = {1806.10293},
  bibsource    = {dblp computer science bibliography, https://dblp.org}
}

@inproceedings{roboagent,
  author       = {Homanga Bharadhwaj and
                  Jay Vakil and
                  Mohit Sharma and
                  Abhinav Gupta and
                  Shubham Tulsiani and
                  Vikash Kumar},
  title        = {RoboAgent: Generalization and Efficiency in Robot Manipulation via
                  Semantic Augmentations and Action Chunking},
  booktitle    = {{IEEE} International Conference on Robotics and Automation, {ICRA}
                  2024, Yokohama, Japan, May 13-17, 2024},
  pages        = {4788--4795},
  publisher    = {{IEEE}},
  year         = {2024},
  url          = {https://doi.org/10.1109/ICRA57147.2024.10611293},
  doi          = {10.1109/ICRA57147.2024.10611293},
  bibsource    = {dblp computer science bibliography, https://dblp.org}
}

@inproceedings{modelsoup,
  title={Model soups: averaging weights of multiple fine-tuned models improves accuracy without increasing inference time},
  author={Wortsman, Mitchell and Ilharco, Gabriel and Gadre, Samir Ya and Roelofs, Rebecca and Gontijo-Lopes, Raphael and Morcos, Ari S and Namkoong, Hongseok and Farhadi, Ali and Carmon, Yair and Kornblith, Simon and others},
  booktitle={International conference on machine learning},
  pages={23965--23998},
  year={2022},
  organization={PMLR}
}

@inproceedings{fishermerging,
  author       = {Michael Matena and
                  Colin Raffel},
  editor       = {Sanmi Koyejo and
                  S. Mohamed and
                  A. Agarwal and
                  Danielle Belgrave and
                  K. Cho and
                  A. Oh},
  title        = {Merging Models with Fisher-Weighted Averaging},
  booktitle    = {Advances in Neural Information Processing Systems 35: Annual Conference
                  on Neural Information Processing Systems 2022, NeurIPS 2022, New Orleans,
                  LA, USA, November 28 - December 9, 2022},
  year         = {2022},
  url          = {http://papers.nips.cc/paper\_files/paper/2022/hash/70c26937fbf3d4600b69a129031b66ec-Abstract-Conference.html},
  bibsource    = {dblp computer science bibliography, https://dblp.org}
}

@inproceedings{regmean,
  author       = {Xisen Jin and
                  Xiang Ren and
                  Daniel Preotiuc{-}Pietro and
                  Pengxiang Cheng},
  title        = {Dataless Knowledge Fusion by Merging Weights of Language Models},
  booktitle    = {The Eleventh International Conference on Learning Representations,
                  {ICLR} 2023, Kigali, Rwanda, May 1-5, 2023},
  publisher    = {OpenReview.net},
  year         = {2023},
  url          = {https://openreview.net/forum?id=FCnohuR6AnM},
  bibsource    = {dblp computer science bibliography, https://dblp.org}
}

@inproceedings{WA1,
  author       = {Jonathan Frankle and
                  Gintare Karolina Dziugaite and
                  Daniel M. Roy and
                  Michael Carbin},
  title        = {Linear Mode Connectivity and the Lottery Ticket Hypothesis},
  booktitle    = {Proceedings of the 37th International Conference on Machine Learning,
                  {ICML} 2020, 13-18 July 2020, Virtual Event},
  series       = {Proceedings of Machine Learning Research},
  volume       = {119},
  pages        = {3259--3269},
  publisher    = {{PMLR}},
  year         = {2020},
  url          = {http://proceedings.mlr.press/v119/frankle20a.html},
  bibsource    = {dblp computer science bibliography, https://dblp.org}
}

@inproceedings{WA2,
  author       = {Behnam Neyshabur and
                  Hanie Sedghi and
                  Chiyuan Zhang},
  editor       = {Hugo Larochelle and
                  Marc'Aurelio Ranzato and
                  Raia Hadsell and
                  Maria{-}Florina Balcan and
                  Hsuan{-}Tien Lin},
  title        = {What is being transferred in transfer learning?},
  booktitle    = {Advances in Neural Information Processing Systems 33: Annual Conference
                  on Neural Information Processing Systems 2020, NeurIPS 2020, December
                  6-12, 2020, virtual},
  year         = {2020},
  url          = {https://proceedings.neurips.cc/paper/2020/hash/0607f4c705595b911a4f3e7a127b44e0-Abstract.html},
  bibsource    = {dblp computer science bibliography, https://dblp.org}
}

@inproceedings{TA,
  author       = {Gabriel Ilharco and
                  Marco T{\'{u}}lio Ribeiro and
                  Mitchell Wortsman and
                  Ludwig Schmidt and
                  Hannaneh Hajishirzi and
                  Ali Farhadi},
  title        = {Editing models with task arithmetic},
  booktitle    = {The Eleventh International Conference on Learning Representations,
                  {ICLR} 2023, Kigali, Rwanda, May 1-5, 2023},
  publisher    = {OpenReview.net},
  year         = {2023},
  url          = {https://openreview.net/forum?id=6t0Kwf8-jrj},
  bibsource    = {dblp computer science bibliography, https://dblp.org}
}

@inproceedings{ties,
  author       = {Prateek Yadav and
                  Derek Tam and
                  Leshem Choshen and
                  Colin A. Raffel and
                  Mohit Bansal},
  editor       = {Alice Oh and
                  Tristan Naumann and
                  Amir Globerson and
                  Kate Saenko and
                  Moritz Hardt and
                  Sergey Levine},
  title        = {TIES-Merging: Resolving Interference When Merging Models},
  booktitle    = {Advances in Neural Information Processing Systems 36: Annual Conference
                  on Neural Information Processing Systems 2023, NeurIPS 2023, New Orleans,
                  LA, USA, December 10 - 16, 2023},
  year         = {2023},
  url          = {http://papers.nips.cc/paper\_files/paper/2023/hash/1644c9af28ab7916874f6fd6228a9bcf-Abstract-Conference.html},
  bibsource    = {dblp computer science bibliography, https://dblp.org}
}

@inproceedings{revla,
  author       = {Sombit Dey and
                  Jan{-}Nico Zaech and
                  Nikolay Nikolov and
                  Luc Van Gool and
                  Danda Pani Paudel},
  title        = {ReVLA: Reverting Visual Domain Limitation of Robotic Foundation Models},
  booktitle    = {{IEEE} International Conference on Robotics and Automation, {ICRA}
                  2025, Atlanta, GA, USA, May 19-23, 2025},
  pages        = {8679--8686},
  publisher    = {{IEEE}},
  year         = {2025},
  url          = {https://doi.org/10.1109/ICRA55743.2025.11128823},
  doi          = {10.1109/ICRA55743.2025.11128823},
  bibsource    = {dblp computer science bibliography, https://dblp.org}
}

@inproceedings{dare,
  author       = {Le Yu and
                  Bowen Yu and
                  Haiyang Yu and
                  Fei Huang and
                  Yongbin Li},
  title        = {Language Models are Super Mario: Absorbing Abilities from Homologous
                  Models as a Free Lunch},
  booktitle    = {Forty-first International Conference on Machine Learning, {ICML} 2024,
                  Vienna, Austria, July 21-27, 2024},
  publisher    = {OpenReview.net},
  year         = {2024},
  url          = {https://openreview.net/forum?id=fq0NaiU8Ex},
  bibsource    = {dblp computer science bibliography, https://dblp.org}
}

@article{do-merging,
  author       = {Shenghe Zheng and
                  Hongzhi Wang and
                  Chenyu Huang and
                  Xiaohui Wang and
                  Tao Chen and
                  Jiayuan Fan and
                  Shuyue Hu and
                  Peng Ye},
  title        = {Decouple and Orthogonalize: {A} Data-Free Framework for LoRA Merging},
  journal      = {CoRR},
  volume       = {abs/2505.15875},
  year         = {2025},
  url          = {https://doi.org/10.48550/arXiv.2505.15875},
  doi          = {10.48550/ARXIV.2505.15875},
  eprinttype    = {arXiv},
  eprint       = {2505.15875},
  bibsource    = {dblp computer science bibliography, https://dblp.org}
}

@inproceedings{breadcrumbs,
  author       = {MohammadReza Davari and
                  Eugene Belilovsky},
  editor       = {Ales Leonardis and
                  Elisa Ricci and
                  Stefan Roth and
                  Olga Russakovsky and
                  Torsten Sattler and
                  G{\"{u}}l Varol},
  title        = {Model Breadcrumbs: Scaling Multi-task Model Merging with Sparse Masks},
  booktitle    = {Computer Vision - {ECCV} 2024 - 18th European Conference, Milan, Italy,
                  September 29-October 4, 2024, Proceedings, Part {LXXV}},
  series       = {Lecture Notes in Computer Science},
  volume       = {15133},
  pages        = {270--287},
  publisher    = {Springer},
  year         = {2024},
  url          = {https://doi.org/10.1007/978-3-031-73226-3\_16},
  doi          = {10.1007/978-3-031-73226-3\_16},
  bibsource    = {dblp computer science bibliography, https://dblp.org}
}

@article{pem-composition,
  author       = {Jinghan Zhang and
                  Shiqi Chen and
                  Junteng Liu and
                  Junxian He},
  title        = {Composing Parameter-Efficient Modules with Arithmetic Operations},
  journal      = {CoRR},
  volume       = {abs/2306.14870},
  year         = {2023},
  url          = {https://doi.org/10.48550/arXiv.2306.14870},
  doi          = {10.48550/ARXIV.2306.14870},
  eprinttype    = {arXiv},
  eprint       = {2306.14870},
  bibsource    = {dblp computer science bibliography, https://dblp.org}
}

@inproceedings{pcb-merging,
  author       = {Guodong Du and
                  Junlin Lee and
                  Jing Li and
                  Runhua Jiang and
                  Yifei Guo and
                  Shuyang Yu and
                  Hanting Liu and
                  Sim Kuan Goh and
                  Ho{-}Kin Tang and
                  Daojing He and
                  Min Zhang},
  editor       = {Amir Globersons and
                  Lester Mackey and
                  Danielle Belgrave and
                  Angela Fan and
                  Ulrich Paquet and
                  Jakub M. Tomczak and
                  Cheng Zhang},
  title        = {Parameter Competition Balancing for Model Merging},
  booktitle    = {Advances in Neural Information Processing Systems 38: Annual Conference
                  on Neural Information Processing Systems 2024, NeurIPS 2024, Vancouver,
                  BC, Canada, December 10 - 15, 2024},
  year         = {2024},
  url          = {http://papers.nips.cc/paper\_files/paper/2024/hash/99fc8bc48b917c301a80cb74d91c0c06-Abstract-Conference.html},
  bibsource    = {dblp computer science bibliography, https://dblp.org}
}

@article{cabs,
  author       = {Zongzhen Yang and
                  Binhang Qi and
                  Hailong Sun and
                  Wenrui Long and
                  Ruobing Zhao and
                  Xiang Gao},
  title        = {{CABS:} Conflict-Aware and Balanced Sparsification for Enhancing Model
                  Merging},
  journal      = {CoRR},
  volume       = {abs/2503.01874},
  year         = {2025},
  url          = {https://doi.org/10.48550/arXiv.2503.01874},
  doi          = {10.48550/ARXIV.2503.01874},
  eprinttype    = {arXiv},
  eprint       = {2503.01874},
  bibsource    = {dblp computer science bibliography, https://dblp.org}
}

@article{wudi,
  author       = {Runxi Cheng and
                  Feng Xiong and
                  Yongxian Wei and
                  Wanyun Zhu and
                  Chun Yuan},
  title        = {Whoever Started the Interference Should End It: Guiding Data-Free
                  Model Merging via Task Vectors},
  journal      = {CoRR},
  volume       = {abs/2503.08099},
  year         = {2025},
  url          = {https://doi.org/10.48550/arXiv.2503.08099},
  doi          = {10.48550/ARXIV.2503.08099},
  eprinttype    = {arXiv},
  eprint       = {2503.08099},
  bibsource    = {dblp computer science bibliography, https://dblp.org}
}

@article{cat-merging,
  author       = {Wenju Sun and
                  Qingyong Li and
                  Yangli{-}ao Geng and
                  Boyang Li},
  title        = {{CAT} Merging: {A} Training-Free Approach for Resolving Conflicts
                  in Model Merging},
  journal      = {CoRR},
  volume       = {abs/2505.06977},
  year         = {2025},
  url          = {https://doi.org/10.48550/arXiv.2505.06977},
  doi          = {10.48550/ARXIV.2505.06977},
  eprinttype    = {arXiv},
  eprint       = {2505.06977},
  bibsource    = {dblp computer science bibliography, https://dblp.org}
}

@inproceedings{tsvm,
  author       = {Antonio Andrea Gargiulo and
                  Donato Crisostomi and
                  Maria Sofia Bucarelli and
                  Simone Scardapane and
                  Fabrizio Silvestri and
                  Emanuele Rodol{\`{a}}},
  title        = {Task Singular Vectors: Reducing Task Interference in Model Merging},
  booktitle    = {{IEEE/CVF} Conference on Computer Vision and Pattern Recognition,
                  {CVPR} 2025, Nashville, TN, USA, June 11-15, 2025},
  pages        = {18695--18705},
  publisher    = {Computer Vision Foundation / {IEEE}},
  year         = {2025},
  url          = {https://openaccess.thecvf.com/content/CVPR2025/html/Gargiulo\_Task\_Singular\_Vectors\_Reducing\_Task\_Interference\_in\_Model\_Merging\_CVPR\_2025\_paper.html},
  doi          = {10.1109/CVPR52734.2025.01742},
  bibsource    = {dblp computer science bibliography, https://dblp.org}
}

@inproceedings{twin-merging,
  author       = {Zhenyi Lu and
                  Chenghao Fan and
                  Wei Wei and
                  Xiaoye Qu and
                  Dangyang Chen and
                  Yu Cheng},
  editor       = {Amir Globersons and
                  Lester Mackey and
                  Danielle Belgrave and
                  Angela Fan and
                  Ulrich Paquet and
                  Jakub M. Tomczak and
                  Cheng Zhang},
  title        = {Twin-Merging: Dynamic Integration of Modular Expertise in Model Merging},
  booktitle    = {Advances in Neural Information Processing Systems 38: Annual Conference
                  on Neural Information Processing Systems 2024, NeurIPS 2024, Vancouver,
                  BC, Canada, December 10 - 15, 2024},
  year         = {2024},
  url          = {http://papers.nips.cc/paper\_files/paper/2024/hash/8fcd17eb91bae20d9826786d7d6be799-Abstract-Conference.html},
  bibsource    = {dblp computer science bibliography, https://dblp.org}
}

@inproceedings{knots,
  author       = {George Stoica and
                  Pratik Ramesh and
                  Boglarka Ecsedi and
                  Leshem Choshen and
                  Judy Hoffman},
  title        = {Model merging with {SVD} to tie the Knots},
  booktitle    = {The Thirteenth International Conference on Learning Representations,
                  {ICLR} 2025, Singapore, April 24-28, 2025},
  publisher    = {OpenReview.net},
  year         = {2025},
  url          = {https://openreview.net/forum?id=67X93aZHII},
  bibsource    = {dblp computer science bibliography, https://dblp.org}
}

@article{iso-cts,
  author       = {Daniel Marczak and
                  Simone Magistri and
                  Sebastian Cygert and
                  Bartlomiej Twardowski and
                  Andrew D. Bagdanov and
                  Joost van de Weijer},
  title        = {No Task Left Behind: Isotropic Model Merging with Common and Task-Specific
                  Subspaces},
  journal      = {CoRR},
  volume       = {abs/2502.04959},
  year         = {2025},
  url          = {https://doi.org/10.48550/arXiv.2502.04959},
  doi          = {10.48550/ARXIV.2502.04959},
  eprinttype    = {arXiv},
  eprint       = {2502.04959},
  bibsource    = {dblp computer science bibliography, https://dblp.org}
}

@misc{robust-merging,
      title={RobustMerge: Parameter-Efficient Model Merging for MLLMs with Direction Robustness}, 
      author={Fanhu Zeng and Haiyang Guo and Fei Zhu and Li Shen and Hao Tang},
      year={2025},
      eprint={2502.17159},
      archivePrefix={arXiv},
      primaryClass={cs.CV},
      url={https://arxiv.org/abs/2502.17159}, 
}

@inproceedings{tall-mask,
  author       = {Ke Wang and
                  Nikolaos Dimitriadis and
                  Guillermo Ortiz{-}Jim{\'{e}}nez and
                  Fran{\c{c}}ois Fleuret and
                  Pascal Frossard},
  title        = {Localizing Task Information for Improved Model Merging and Compression},
  booktitle    = {Forty-first International Conference on Machine Learning, {ICML} 2024,
                  Vienna, Austria, July 21-27, 2024},
  publisher    = {OpenReview.net},
  year         = {2024},
  url          = {https://openreview.net/forum?id=DWT9uiGjxT},
  bibsource    = {dblp computer science bibliography, https://dblp.org}
}

@article{smile,
  author       = {Anke Tang and
                  Li Shen and
                  Yong Luo and
                  Shuai Xie and
                  Han Hu and
                  Lefei Zhang and
                  Bo Du and
                  Dacheng Tao},
  title        = {{SMILE:} Zero-Shot Sparse Mixture of Low-Rank Experts Construction
                  From Pre-Trained Foundation Models},
  journal      = {CoRR},
  volume       = {abs/2408.10174},
  year         = {2024},
  url          = {https://doi.org/10.48550/arXiv.2408.10174},
  doi          = {10.48550/ARXIV.2408.10174},
  eprinttype    = {arXiv},
  eprint       = {2408.10174},
  bibsource    = {dblp computer science bibliography, https://dblp.org}
}

@article{calm-merging,
  author       = {Kunda Yan and
                  Min Zhang and
                  Sen Cui and
                  Zikun Qu and
                  Bo Jiang and
                  Feng Liu and
                  Changshui Zhang},
  title        = {{CALM:} Consensus-Aware Localized Merging for Multi-Task Learning},
  journal      = {CoRR},
  volume       = {abs/2506.13406},
  year         = {2025},
  url          = {https://doi.org/10.48550/arXiv.2506.13406},
  doi          = {10.48550/ARXIV.2506.13406},
  eprinttype    = {arXiv},
  eprint       = {2506.13406},
  bibsource    = {dblp computer science bibliography, https://dblp.org}
}

@inproceedings{emr-merging,
  author       = {Chenyu Huang and
                  Peng Ye and
                  Tao Chen and
                  Tong He and
                  Xiangyu Yue and
                  Wanli Ouyang},
  editor       = {Amir Globersons and
                  Lester Mackey and
                  Danielle Belgrave and
                  Angela Fan and
                  Ulrich Paquet and
                  Jakub M. Tomczak and
                  Cheng Zhang},
  title        = {EMR-Merging: Tuning-Free High-Performance Model Merging},
  booktitle    = {Advances in Neural Information Processing Systems 38: Annual Conference
                  on Neural Information Processing Systems 2024, NeurIPS 2024, Vancouver,
                  BC, Canada, December 10 - 15, 2024},
  year         = {2024},
  url          = {http://papers.nips.cc/paper\_files/paper/2024/hash/dda5cac5272a9bcd4bc73d90bc725ef1-Abstract-Conference.html},
  bibsource    = {dblp computer science bibliography, https://dblp.org}
}

@article{openvla-oft,
  author       = {Moo Jin Kim and
                  Chelsea Finn and
                  Percy Liang},
  title        = {Fine-Tuning Vision-Language-Action Models: Optimizing Speed and Success},
  journal      = {CoRR},
  volume       = {abs/2502.19645},
  year         = {2025},
  url          = {https://doi.org/10.48550/arXiv.2502.19645},
  doi          = {10.48550/ARXIV.2502.19645},
  eprinttype    = {arXiv},
  eprint       = {2502.19645},
  bibsource    = {dblp computer science bibliography, https://dblp.org}
}

@misc{qwen25,
      title={Qwen2.5 Technical Report}, 
      author={An Yang and Baosong Yang and Beichen Zhang and Binyuan Hui and Bo Zheng and Bowen Yu and Chengyuan Li and Dayiheng Liu and Fei Huang and Haoran Wei and Huan Lin and Jian Yang and Jianhong Tu and Jianwei Zhang and Jianxin Yang and Jiaxi Yang and Jingren Zhou and Junyang Lin and Kai Dang and Keming Lu and Keqin Bao and Kexin Yang and Le Yu and Mei Li and Mingfeng Xue and Pei Zhang and Qin Zhu and Rui Men and Runji Lin and Tianhao Li and Tianyi Tang and Tingyu Xia and Xingzhang Ren and Xuancheng Ren and Yang Fan and Yang Su and Yichang Zhang and Yu Wan and Yuqiong Liu and Zeyu Cui and Zhenru Zhang and Zihan Qiu},
      year={2025},
      eprint={2412.15115},
      archivePrefix={arXiv},
      primaryClass={cs.CL},
      url={https://arxiv.org/abs/2412.15115}, 
}

@article{gr00t,
  author       = {Johan Bjorck and
                  Fernando Casta{\~{n}}eda and
                  Nikita Cherniadev and
                  Xingye Da and
                  Runyu Ding and
                  Linxi Fan and
                  Yu Fang and
                  Dieter Fox and
                  Fengyuan Hu and
                  Spencer Huang and
                  Joel Jang and
                  Zhenyu Jiang and
                  Jan Kautz and
                  Kaushil Kundalia and
                  Lawrence Lao and
                  Zhiqi Li and
                  Zongyu Lin and
                  Kevin Lin and
                  Guilin Liu and
                  Edith LLontop and
                  Loic Magne and
                  Ajay Mandlekar and
                  Avnish Narayan and
                  Soroush Nasiriany and
                  Scott Reed and
                  You Liang Tan and
                  Guanzhi Wang and
                  Zu Wang and
                  Jing Wang and
                  Qi Wang and
                  Jiannan Xiang and
                  Yuqi Xie and
                  Yinzhen Xu and
                  Zhenjia Xu and
                  Seonghyeon Ye and
                  Zhiding Yu and
                  Ao Zhang and
                  Hao Zhang and
                  Yizhou Zhao and
                  Ruijie Zheng and
                  Yuke Zhu},
  title        = {{GR00T} {N1:} An Open Foundation Model for Generalist Humanoid Robots},
  journal      = {CoRR},
  volume       = {abs/2503.14734},
  year         = {2025},
  url          = {https://doi.org/10.48550/arXiv.2503.14734},
  doi          = {10.48550/ARXIV.2503.14734},
  eprinttype    = {arXiv},
  eprint       = {2503.14734},
  bibsource    = {dblp computer science bibliography, https://dblp.org}
}

@article{spatialvla,
  author       = {Delin Qu and
                  Haoming Song and
                  Qizhi Chen and
                  Yuanqi Yao and
                  Xinyi Ye and
                  Yan Ding and
                  Zhigang Wang and
                  JiaYuan Gu and
                  Bin Zhao and
                  Dong Wang and
                  Xuelong Li},
  title        = {SpatialVLA: Exploring Spatial Representations for Visual-Language-Action
                  Model},
  journal      = {CoRR},
  volume       = {abs/2501.15830},
  year         = {2025},
  url          = {https://doi.org/10.48550/arXiv.2501.15830},
  doi          = {10.48550/ARXIV.2501.15830},
  eprinttype    = {arXiv},
  eprint       = {2501.15830},
  bibsource    = {dblp computer science bibliography, https://dblp.org}
}

@article{smolvla,
  author       = {Mustafa Shukor and
                  Dana Aubakirova and
                  Francesco Capuano and
                  Pepijn Kooijmans and
                  Steven Palma and
                  Adil Zouitine and
                  Michel Aractingi and
                  Caroline Pascal and
                  Martino Russi and
                  Andr{\'{e}}s Marafioti and
                  Simon Alibert and
                  Matthieu Cord and
                  Thomas Wolf and
                  R{\'{e}}mi Cad{\`{e}}ne},
  title        = {SmolVLA: {A} Vision-Language-Action Model for Affordable and Efficient
                  Robotics},
  journal      = {CoRR},
  volume       = {abs/2506.01844},
  year         = {2025},
  url          = {https://doi.org/10.48550/arXiv.2506.01844},
  doi          = {10.48550/ARXIV.2506.01844},
  eprinttype    = {arXiv},
  eprint       = {2506.01844},
  bibsource    = {dblp computer science bibliography, https://dblp.org}
}

@inproceedings{octo,
  author       = {Dibya Ghosh and
                  Homer Rich Walke and
                  Karl Pertsch and
                  Kevin Black and
                  Oier Mees and
                  Sudeep Dasari and
                  Joey Hejna and
                  Tobias Kreiman and
                  Charles Xu and
                  Jianlan Luo and
                  You Liang Tan and
                  Lawrence Yunliang Chen and
                  Quan Vuong and
                  Ted Xiao and
                  Pannag R. Sanketi and
                  Dorsa Sadigh and
                  Chelsea Finn and
                  Sergey Levine},
  editor       = {Dana Kulic and
                  Gentiane Venture and
                  Kostas E. Bekris and
                  Enrique Coronado},
  title        = {Octo: An Open-Source Generalist Robot Policy},
  booktitle    = {Robotics: Science and Systems XX, Delft, The Netherlands, July 15-19,
                  2024},
  year         = {2024},
  url          = {https://doi.org/10.15607/RSS.2024.XX.090},
  doi          = {10.15607/RSS.2024.XX.090},
  bibsource    = {dblp computer science bibliography, https://dblp.org}
}

@article{tinyvla,
  author       = {Junjie Wen and
                  Yichen Zhu and
                  Jinming Li and
                  Minjie Zhu and
                  Kun Wu and
                  Zhiyuan Xu and
                  Ning Liu and
                  Ran Cheng and
                  Chaomin Shen and
                  Yaxin Peng and
                  Feifei Feng and
                  Jian Tang},
  title        = {TinyVLA: Towards Fast, Data-Efficient Vision-Language-Action Models
                  for Robotic Manipulation},
  journal      = {CoRR},
  volume       = {abs/2409.12514},
  year         = {2024},
  url          = {https://doi.org/10.48550/arXiv.2409.12514},
  doi          = {10.48550/ARXIV.2409.12514},
  eprinttype    = {arXiv},
  eprint       = {2409.12514},
  bibsource    = {dblp computer science bibliography, https://dblp.org}
}

@inproceedings{magma,
  author       = {Jianwei Yang and
                  Reuben Tan and
                  Qianhui Wu and
                  Ruijie Zheng and
                  Baolin Peng and
                  Yongyuan Liang and
                  Yu Gu and
                  Mu Cai and
                  Seonghyeon Ye and
                  Joel Jang and
                  Yuquan Deng and
                  Jianfeng Gao},
  title        = {Magma: {A} Foundation Model for Multimodal {AI} Agents},
  booktitle    = {{IEEE/CVF} Conference on Computer Vision and Pattern Recognition,
                  {CVPR} 2025, Nashville, TN, USA, June 11-15, 2025},
  pages        = {14203--14214},
  publisher    = {Computer Vision Foundation / {IEEE}},
  year         = {2025},
  url          = {https://openaccess.thecvf.com/content/CVPR2025/html/Yang\_Magma\_A\_Foundation\_Model\_for\_Multimodal\_AI\_Agents\_CVPR\_2025\_paper.html},
  doi          = {10.1109/CVPR52734.2025.01325},
  bibsource    = {dblp computer science bibliography, https://dblp.org}
}

@article{vlm2vla,
  author       = {Asher J. Hancock and
                  Xindi Wu and
                  Lihan Zha and
                  Olga Russakovsky and
                  Anirudha Majumdar},
  title        = {Actions as Language: Fine-Tuning VLMs into VLAs Without Catastrophic
                  Forgetting},
  journal      = {CoRR},
  volume       = {abs/2509.22195},
  year         = {2025},
  url          = {https://doi.org/10.48550/arXiv.2509.22195},
  doi          = {10.48550/ARXIV.2509.22195},
  eprinttype    = {arXiv},
  eprint       = {2509.22195},
  bibsource    = {dblp computer science bibliography, https://dblp.org}
}

@inproceedings{uq_merge,
  author       = {Huaizhi Qu and
                  Xinyu Zhao and
                  Jie Peng and
                  Kwonjoon Lee and
                  Behzad Dariush and
                  Tianlong Chen},
  editor       = {Wanxiang Che and
                  Joyce Nabende and
                  Ekaterina Shutova and
                  Mohammad Taher Pilehvar},
  title        = {UQ-Merge: Uncertainty Guided Multimodal Large Language Model Merging},
  booktitle    = {Findings of the Association for Computational Linguistics, {ACL} 2025,
                  Vienna, Austria, July 27 - August 1, 2025},
  pages        = {1401--1417},
  publisher    = {Association for Computational Linguistics},
  year         = {2025},
  url          = {https://aclanthology.org/2025.findings-acl.73/},
  bibsource    = {dblp computer science bibliography, https://dblp.org}
}

@article{bring_r_to_v,
  author       = {Shiqi Chen and
                  Jinghan Zhang and
                  Tongyao Zhu and
                  Wei Liu and
                  Siyang Gao and
                  Miao Xiong and
                  Manling Li and
                  Junxian He},
  title        = {Bring Reason to Vision: Understanding Perception and Reasoning through
                  Model Merging},
  journal      = {CoRR},
  volume       = {abs/2505.05464},
  year         = {2025},
  url          = {https://doi.org/10.48550/arXiv.2505.05464},
  doi          = {10.48550/ARXIV.2505.05464},
  eprinttype    = {arXiv},
  eprint       = {2505.05464},
  bibsource    = {dblp computer science bibliography, https://dblp.org}
}

@inproceedings{adamms,
  author       = {Yiyang Du and
                  Xiaochen Wang and
                  Chi Chen and
                  Jiabo Ye and
                  Yiru Wang and
                  Peng Li and
                  Ming Yan and
                  Ji Zhang and
                  Fei Huang and
                  Zhifang Sui and
                  Maosong Sun and
                  Yang Liu},
  title        = {AdaMMS: Model Merging for Heterogeneous Multimodal Large Language
                  Models with Unsupervised Coefficient Optimization},
  booktitle    = {{IEEE/CVF} Conference on Computer Vision and Pattern Recognition,
                  {CVPR} 2025, Nashville, TN, USA, June 11-15, 2025},
  pages        = {9413--9422},
  publisher    = {Computer Vision Foundation / {IEEE}},
  year         = {2025},
  url          = {https://openaccess.thecvf.com/content/CVPR2025/html/Du\_AdaMMS\_Model\_Merging\_for\_Heterogeneous\_Multimodal\_Large\_Language\_Models\_with\_CVPR\_2025\_paper.html},
  doi          = {10.1109/CVPR52734.2025.00879},
  bibsource    = {dblp computer science bibliography, https://dblp.org}
}
}

\clearpage
\setcounter{page}{1}
\maketitlesupplementary

\section{Experimental Details}
We summarize all fine-tuning hyperparameters used across LIBERO, RoboTwin, and real-world SO-101 experiments in Table~\ref{tab:exp_details}. All experiments use the same VLM backbone and training configuration unless otherwise specified.

\begin{table}[h]
\centering
\caption{\textbf{Fine-tuning hyperparameters used in all experiments.}}
\label{tab:exp_details}
\begin{tabular}{lc}
\toprule
\textbf{Hyperparameter} & \textbf{Value} \\
\midrule
Backbone & Qwen-2.5 (0.5B) \\
Batch size & 8 \\
Learning rate & $5\times10^{-4}$ \\
LoRA rank & 32 \\
Use proprioception & True \\
Num images & 2 (3 for Robotwin) \\
Gradient step & 30k (50k for LIBERO-Long) \\
\bottomrule
\end{tabular}
\end{table}

\section{Algorithm Details}
In this section, we give a detailed algorithm description in Algorithm.~\ref{alg:mergevla_routing} of how MergeVLA performs inference using our test-time task router when the task identity is unknown.

\begin{algorithm}[ht]
\caption{Test-time Task Routing and Inference in MergeVLA}
\label{alg:mergevla_routing}
    \begin{algorithmic}[1]
    \State \textbf{Inputs:} Task masks $\{\mathbf{S}_m\}_{m=1}^{M}$; Expert heads $\{\mathbf{H}_m^{\,l\rightarrow L}\}_{m=1}^{M}$; Pretrained VLM weights $\Theta_0$; Merged task vector $\tau_{\mathrm{merge}}$; Value projections of the action expert at block $l$: $\mathbf{V}_{\mathrm{T}, m}^{\,l},\;\mathbf{V}_{\mathrm{A}, m}^{\,l}$; Initial observation $(\mathbf{I}_0^{v}, \mathbf{I}_0^{w}, L)$
    
    \State \textbf{Routing phase (at $t=0$):}
    \For{$m=1$ to $M$}
        \State $\Theta_{\mathrm{VLM}}^{(m)} = \Theta_0 + \mathbf{S}_m \odot \tau_{\mathrm{merge}}$ \Comment{Construct masked VLM}
        \State $\mathbf{h}_{\mathrm{T},m}^{\,l},\; \mathbf{h}_{\mathrm{A},m}^{\,l} = \Theta_{\mathrm{VLM}}^{(m)}(\mathbf{I}_0^{v}, \mathbf{I}_0^{w}, L)$ \Comment{Extract $l$-th block hidden states}
        \State $\mathbf{V}_{\mathrm{T}, m}^{\,l} = \mathbf{L}_{\mathrm{T}, m}^{\,l}\,\mathbf{\Sigma}_{\mathrm{T}, m}^{\,l}\,(\mathbf{R}_{\mathrm{T}, m}^{\,l})^\top$
        \State $\mathbf{V}_{\mathrm{A}, m}^{\,l} = \mathbf{L}_{\mathrm{A}, m}^{\,l}\,\mathbf{\Sigma}_{\mathrm{A}, m}^{\,l}\,(\mathbf{R}_{\mathrm{A}, m}^{\,l})^\top$
        \State $\mathbf{r}_{\mathrm{T},m} = \big\| \mathbf{P}_{\mathrm{T}, m}^{\,l} \mathbf{h}^{\,l}_{\mathrm{A},m} \big\|_2$ \Comment{Choose top-$r_k$ singular vectors from $\mathbf{R}$ to get $\mathbf{P}$}
        \State $\mathbf{r}_{\mathrm{A},m} = \big\| \mathbf{P}_{\mathrm{A}, m}^{\,l} \mathbf{h}^{\,l}_{\mathrm{T},m} \big\|_2$
        \State $\mathbf{r}_m = \tfrac{1}{2}\big(\mathbf{r}_{\mathrm{T},m} + \mathbf{r}_{\mathrm{A},m}\big)$
    \EndFor
    \State $m^{*} = \arg\max_m \operatorname{softmax}(\mathbf{r}).$ \Comment{Normalize scores with softmax and select task index}
    \State \Return{$m^{*}$} 
    \State \textbf{Inference phase:}
    Use $\mathbf{S}_{m^{*}}$, and expert head $\mathbf{H}_{m^{*}}^{\,l\rightarrow L}$ for all $t\ge 0$.
    \end{algorithmic}
\end{algorithm}

\section{Preliminary Investigation on OpenVLA}
In the early stage of this work, we explored the feasibility of directly applying existing model merging methods to OpenVLA~\cite{OpenVLA}, a popular VLA model. OpenVLA consists of three main components: a vision backbone, a projector, and a language model. The language model itself contains 32 transformer blocks followed by a single-layer MLP head (lm\_head). We first attempted to merge all components of OpenVLA using Weighted Average and Task Arithmetic methods. However, the merged checkpoint completely failed on all tasks. This was surprising, as OpenVLA is essentially a VLM, and previous studies~\cite{adamms, bring_r_to_v, uq_merge} have shown that VLMs can usually be merged successfully. This prompted us to investigate which part of OpenVLA prevents successful merging.

\subsection{Non-mergeable Components in OpenVLA}
To locate the source of failure, we decomposed the model into four submodules:
$\mathrm{A.}$ the vision backbone;
$\mathrm{B.}$ the projector;
$\mathrm{C.}$ the language model body (excluding lm\_head); and
$\mathrm{D.}$ the lm\_head.
We then merged each submodule separately across the four official LIBERO task checkpoints using the existing merging method Iso-CTS~\cite{iso-cts}. Each merged checkpoint was evaluated on 50 trials per subtask. The results are summarized in Table~\ref{tab:openvla_component_merge}, where the gray-highlighted row denotes the single-task fine-tuning performance. From the table, we observe that merging modules $\mathrm{A}$, $\mathrm{B}$, or $\mathrm{D}$ only slightly decreases success rates, whereas merging the language model body ($\mathrm{C}$) causes complete failure on all tasks. This clearly indicates that the language model is the primary source of merging failure.

We hypothesize that this phenomenon arises because VLA tasks impose much stricter precision requirements on the model outputs than typical LLM or VLM tasks. In LLMs or VLMs, outputs are often discrete token sequences or probability distributions, where small deviations are tolerable. In contrast, robotic control requires continuous numeric outputs, where even minor errors can cause irreversible physical or simulated state changes. Once the environment diverges from the model’s training distribution, subsequent actions fail catastrophically. As component $\mathrm{C}$ is directly responsible for decoding actions, it likely accumulates task-specific differences that make naive merging infeasible. This also explains why, as shown in the main paper, applying task masks to preserve localized task information can effectively mitigate such conflicts and enable multi-task unification.

Additional patterns can be observed from Table~\ref{tab:openvla_component_merge}. Interestingly, merging only the vision backbone ($\mathrm{A}$) consistently yields higher success rates than merging both A and B together. This counterintuitive result suggests that, in robotic domains, all modules may exhibit nontrivial task interference, and increasing the number of merged modules amplifies this conflict. In contrast, merging the lm\_head ($\mathrm{D}$) has little impact on performance. It is only about 3 points below the fine-tuned baseline. Moreover, combinations such as $\mathrm{A+D}$ and $\mathrm{A+B+D}$ show negligible difference from $\mathrm{A}$ and $\mathrm{A+B}$ respectively. To further confirm this, we swapped the lm\_head ($\mathrm{D}$) between the LIBERO-Object and LIBERO-Spatial tasks and tested the model on LIBERO-Spatial. Remarkably, it still achieved an 82\% success rate over 20 trials, indicating that heads of OpenVLA are largely interchangeable across tasks.

\begin{table}[!htbp]
    \small
    \centering
    \caption{Success rates on the four LIBERO task suites when merging different components of OpenVLA~\cite{OpenVLA} using the Iso-CTS~\cite{iso-cts} merging method. Each merged checkpoint combines four task-specific models, while unmerged components retain their original weights. During evaluation, each subtask is tested with 50 trials. \textcolor{gray}{Gray-highlighted} row indicates the success rates of individually fine-tuned models.}
    \label{tab:openvla_component_merge}
    \begin{tabular}{lccccc}
        \toprule
        \textbf{Method} & \textbf{Spatial} & \textbf{Object} & \textbf{Goal} & \textbf{Long} & \textbf{Avg.} \\
        \midrule
        \rowcolor{highlightgray}
        Finetuned& 84.7& 88.4& 79.2& 53.7& 76.5\\
        $\mathrm{A}$& 61.4& 60.8& 59.0& 8.4& 47.4\\
        $\mathrm{A+B}$&  56.6&  58.0&  55.6&  6.6&  44.2\\
        $\mathrm{C}$&  0.0&  0.0&  0.0&  0.0&  0.0\\
        $\mathrm{D}$& 83.4& 88.8& 72.6& 49.6&  73.6\\
        $\mathrm{A+D}$& 61.0& 61.0& 62.6&8.4 &48.3  \\
        $\mathrm{A+B+D}$& 58.0& 57.4& 53.8&7.4 &44.2\\
        \bottomrule
    \end{tabular}
\end{table}

\subsection{Progressive Block-wise Merging of the Language Model}

\begin{figure}[!htbp]
    \centering
    \includegraphics[width=0.3\textwidth]{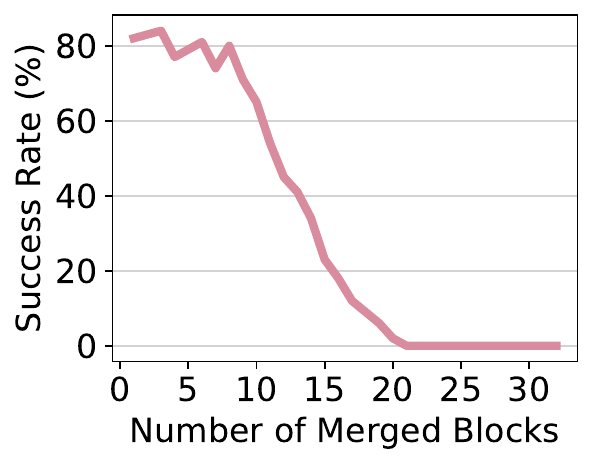}
    \caption{Success rate on the LIBERO-Spatial task when progressively merging the first $k$ language model blocks of OpenVLA~\cite{OpenVLA} using the Iso-CTS~\cite{iso-cts} merging algorithm. Each configuration merges four task-specific checkpoints and is evaluated over 10 trials per subtask.}
    \label{fig:num_merged_block}
\end{figure}

To further analyze why the language model component cannot be merged, we conducted a block-wise study by progressively merging the first $k$ transformer blocks (from 1 to 32) while keeping all other parts fixed to the LIBERO-Spatial task weights. Each merged model was evaluated on 10 trials per subtask, and results are shown in Figure~\ref{fig:num_merged_block}. When merging only a few shallow blocks (\textit{e.g.,} up to 8), the model still achieved roughly 80\% success rate. However, as the number of merged blocks increased, performance degraded sharply, and beyond 21 merged blocks the model completely failed. This again validates our hypothesis: Task conflicts grow with layer depth, and deeper layers show stronger task-specific divergence that hinders effective merging.

\begin{figure}[!htbp]
    \centering
    \includegraphics[width=0.3\textwidth]{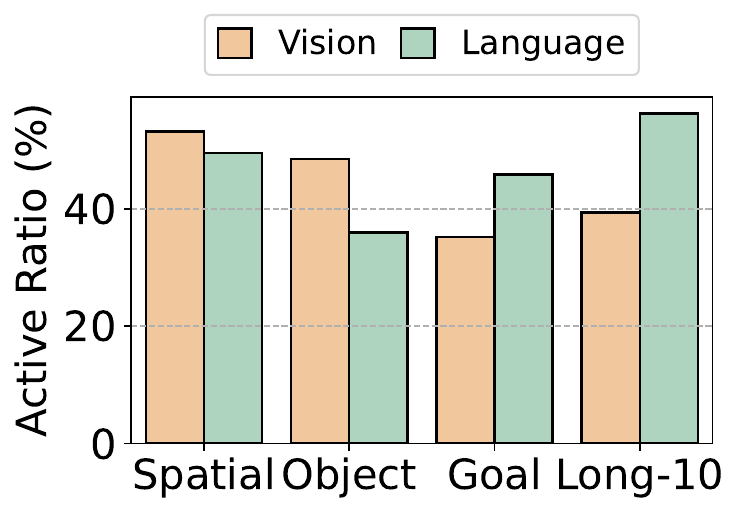}
    \caption{Mask active ratios for each LIBERO task suite, computed for both the vision backbone and the language model components following the same definition as in the main paper. Masks are obtained using the Task Arithmetic~\cite{TA} merging method with $\lambda=0.6$.}
    \label{fig:vis mask ratio VL}
\end{figure}

\section{Visualization of Mask Ratios in the Vision Backbone and Language Model}
To examine how task masks behave across different components of the VLM, we visualize the mask active ratio for each LIBERO task suite. The mask active ratio measures the proportion of positions where the task mask is active (\textit{i.e.,} set to True), indicating that the model uses the pretrained weight + task vector at that location. In contrast, inactive positions fall back to the pretrained weights only. Because the VLM consists of a vision backbone and a language model, we compute the active ratio separately for these two parts to analyze their task-specific behavior. A higher active ratio suggests stronger task-specific contributions, while a lower ratio indicates greater reliance on pretrained weights. Figure~\ref{fig:vis mask ratio VL} shows that the patterns across tasks differ markedly between the vision backbone and the language model. For example, LIBERO-Long exhibits very low activation in the vision backbone but the highest activation in the language model, whereas LIBERO-Object shows the opposite trend—high activation in the vision backbone but minimal activation in the language model. LIBERO-Spatial, in contrast, maintains relatively high and balanced activation across both components. These observations suggest that visual and linguistic pathways contribute task-specific information in distinct ways, offering useful insights for future work on understanding and leveraging task specialization in VLA models.

\end{document}